\pdfoutput=1

\documentclass[11pt]{article}

\usepackage[final]{acl}

\usepackage{times}
\usepackage{latexsym}
\usepackage{graphicx}      
\usepackage{subcaption}    
\usepackage[T1]{fontenc}
\usepackage{enumitem}
\usepackage[utf8]{inputenc}
\usepackage{kotex} 
\usepackage{subcaption}    
\usepackage{microtype}

\usepackage{inconsolata}
\usepackage{multirow}
\usepackage{colortbl}
\usepackage{booktabs}
\usepackage{graphicx}
\usepackage{comment}

\usepackage{soul}
\usepackage{hyperref}
\usepackage{tabularx}
\usepackage{amssymb}
\usepackage{tikz}
\usepackage{pgfplots}
\usepackage{filecontents}
\usepackage{array}
\usepackage{caption}
\usepackage{amsmath}
\usepackage{xcolor}
\usepackage{listings}
\usepackage{float}
\usepackage{enumitem}
\usepackage{textcomp}
\pgfplotsset{compat=1.18}
\title{Identifying and Mitigating Bottlenecks in Role-Playing Agents:\\ A Systematic Study of Disentangling Character Profile Axes}

\author{
\textbf{Yonghyun Jun}\thanks{\;\;Equal contribution.} \quad
\textbf{Junhyuk Choi}\footnotemark[1] \quad
\textbf{Jeonghyun Park} \quad
\textbf{Jihyeong Park} \\
\textbf{Liu Nicole Geumheon} \quad
\textbf{Hwanhee Lee}\thanks{\;\;Corresponding author.} \\
Chung-Ang University, Seoul, Korea \\
\texttt{\{zgold5670,chlwnsgur129,tom0365,g2hyeong,febygh,hwanheelee\}@cau.ac.kr} \\
\url{https://github.com/yonghyun010102/RPA_Bottleneck.git}
}

\begin{document}
\maketitle


\begin{abstract}
While Large Language Model (LLM) role-playing agents have advanced rapidly, it remains unclear which profile elements genuinely drive role-playing quality.
To bridge this gap, we introduce a systematic diagnostic framework that disentangles the impact of character profiles along three axes: \textbf{Familiarity} (Known vs. Unknown), \textbf{Structure} (Structured vs. Unstructured), and \textbf{Disposition} (Moral vs. Immoral).  
Utilizing a unified hierarchical schema (5 dimensions, 28 fields), we construct a controlled dataset of 211 personas and evaluate five LLMs on both single- and multi-turn interactions.
Our results reveal a striking asymmetry: \textbf{Familiarity} and \textbf{Structure} show negligible impact, while \textbf{Disposition} produces large, consistent performance degradation for immoral characters across all conditions. Further analyses suggest that the Moral--Immoral gap is amplified by post-SFT alignment, and that this degradation varies substantially across profile attributes.
To mitigate this bottleneck, we propose Field-Aware Contrastive Decoding (FACD), a training-free strategy that amplifies suppressed disposition-sensitive signals, significantly closing the performance gap without sacrificing moral-character performance.

\end{abstract}

\section{Introduction}
Advancements in Large Language Models (LLMs) have catalyzed a surge of interest in Role-Playing Agents (RPAs) \citep{10.1145/3586183.3606763,chuang2024simulating,wang-etal-2024-incharacter,li2023chatharuhi,wang2024voyager}. 
Alongside rapid industrial adoption by character-based chatting services such as Character.AI\footnote{https://character.ai/}, extensive research efforts have focused on improving RPA performance through various approaches, including reasoning \citep{chen2024agentverse}, retrieval-augmented generation \citep{huang2024emotional}, and fine-tuning \citep{sun2025identity}.

\begin{figure}
  \centering
  \includegraphics[width=\columnwidth]{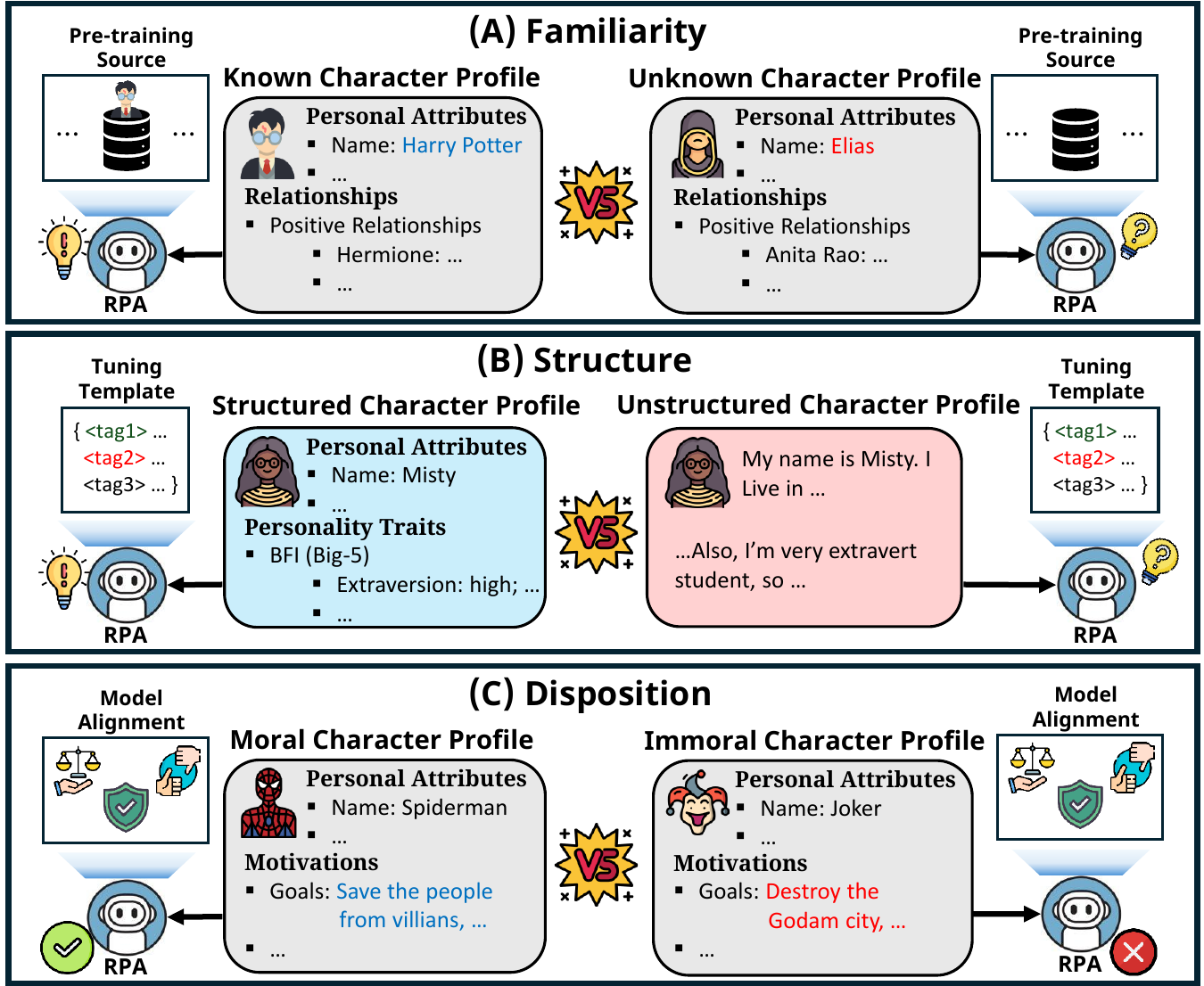}
  \vspace{-5mm}
  \caption{The three axes of character profile: (A) \textbf{Familiarity}: \textit{Known} vs. \textit{Unknown}, (B) \textbf{Structure}: \textit{Structured} vs. \textit{Unstructured}, and  (C) \textbf{Disposition}: \textit{Moral} vs. \textit{Immoral} characters.} 
  \label{fig:motivation}
  \vspace{-6mm}
\end{figure}

Central to these systems is the \textit{character profile}, the foundational specification dictating the portrayed persona. RPA performance can fluctuate significantly based on \textit{\textbf{what}} attributes are selected and \textit{\textbf{how}} they are presented as a prompt. However, existing works fail to systematically disentangle these factors under controlled conditions; they exclusively target either \textit{Known} \citep{wang2025characterbox, liu2024roleagent} or \textit{Unknown} characters \citep{wang2025opencharacter, zhou2025characterbench}, employ ad-hoc templates, and rarely isolate individual traits. Consequently, it remains unclear whether an agent's fidelity is driven by a character's intrinsic nature or its representational format.

In this paper, we conduct a systematic diagnostic study to determine how specific attribute configurations and representational methods drive role-playing performance. 
To ground our analysis in LLM development mechanisms, we hypothesize that profile impacts can be categorized into three axes (Figure \ref{fig:motivation}), each conceptually anchored to a distinct phase of the standard three-stage LLM training pipeline: pre-training, instruction tuning, and post-SFT alignment~\cite{zhao2023survey}.
Specifically, the \textbf{Familiarity} axis (Figure \ref{fig:motivation}(A)), rooted in the \textit{pre-training} phase, distinguishes between characters \textit{Known} to the LLM---whose traits are encoded in its parametric weights---and those that are \textit{Unknown}~\cite{lu-etal-2024-large, sadeq-etal-2024-mitigating} (\textit{\textbf{what}} knowledge). The \textbf{Structure} axis (Figure \ref{fig:motivation}(B)), grounded in \textit{instruction tuning}, contrasts \textit{Structured} schemas against \textit{Unstructured} narratives to probe the model’s structural inductive biases~\cite{sclar2024quantifying, he2024does} (\textit{\textbf{how}} expressed). Finally, the \textbf{Disposition} axis (Figure \ref{fig:motivation}(C)), linked to \textit{post-SFT alignment}~\cite{bai2022training, rafailov2023direct}, concerns the moral polarity of attributes, testing how \textit{Immoral} personas stress the behavioral boundaries established during alignment (\textit{\textbf{what}} nature).

To investigate these three axes under comparable conditions, we design a unified hierarchical profile schema, informed by prior psychological evaluation of LLM personas~\cite{huang2024on}. The schema organizes character information into 5 top-level dimensions and 28 leaf fields, providing a standardized format for profile-level comparisons and a basis for attribute-level analyses. Leveraging this schema, we generate a controlled dataset of 211 characters systematically varying along the \textbf{Familiarity}, \textbf{Structure}, and \textbf{Disposition} axes. We then evaluate five recent open-source LLMs on single-turn~\cite{samuel2024personagym} and multi-turn~\cite{wang2025coser} benchmarks.
The following two research questions guide our investigation:
\vspace{-2mm}
\begin{itemize}[leftmargin=*,itemsep=0.1em]
    \item \textit{RQ1: Which axes of character profile configuration are the primary bottleneck in RPA systems?}
    \item \textit{RQ2: If certain axes introduce performance bottlenecks, how can we mitigate their effects?}
\end{itemize}
\vspace{-2mm}

Our findings reveal a striking asymmetry among the three axes: \textbf{Familiarity} and \textbf{Structure} show negligible impact, whereas \textbf{Disposition} emerges as the dominant driver of role-playing quality. Across all models and benchmarks, \textit{Immoral} characters suffer substantial, consistent performance degradation. 
Counterfactual rewrites further validate this by proving that performance follows dispositional content rather than fixed identity; meanwhile, alignment-stage ablations reveal that post-SFT alignment widens this \textit{Moral}--\textit{Immoral} gap.
We localize this degradation at the field level, finding sharper declines in value-laden motivation dimensions than in stable personality attributes.

To overcome this, we propose \textit{Field-Aware Contrastive Decoding (FACD)}, a training-free decoding strategy that selectively amplifies these suppressed behavioral signals. We show that FACD significantly narrows the performance gap between moral and immoral characters. Ultimately, our insights and FACD offer a practical blueprint for deploying a diverse cast of characters, including complex antagonistic figures.
\section{Related Work}
\subsection{Role-playing Agents}
Recent RPA advancements are driven by new benchmarks and frameworks focusing on historical or fictional figures~\citep{wang2024rolellm, liu2024roleagent, ran2025bookworld}. Specialized models align LLMs with specific roles~\citep{zhou2024characterglm, shao-etal-2023-character, wang2025coser}, evaluated through interview or interactive settings~\citep{wang-etal-2024-incharacter, wang2025characterbox, samuel2024personagym}.
Research on known characters highlights the dual nature of LLM parametric knowledge: while it can trigger character hallucination~\citep{sadeq-etal-2024-mitigating, ahn-etal-2024-timechara, zhang-etal-2025-revealing}, it also enables self-aligned role-play without external metadata~\citep{lu-etal-2024-large}. 

Beyond known figures, growing interest targets ``unknown characters'' relying on user-defined personas. Frameworks for training customizable, synthetic personas have emerged~\citep{wang2025opencharacter, yang-etal-2025-crafting}, alongside benchmarks evaluating customization fidelity~\citep{zhou2025characterbench}. Because existing studies rely on ad-hoc, unstandardized templates, the performance disparity between \textit{Known} and \textit{Unknown} characters remains unexplored.
We unify Known and Unknown characters under one schema to compare them directly.
 
\subsection{Attribute-level Analysis of RPAs}
Research on persona injection in LLMs has shown that prompt-based persona descriptions can reliably induce corresponding behavioral patterns. \citet{jiang2024personallm} demonstrated that BFI-based \citep{john1991big} personality traits are consistently expressed across both psychometric assessments and dialogue tasks. Subsequent work has examined how different persona specifications affect role-playing quality, finding that behavioral guidelines outperform descriptive sketches \citep{wang2025characterbox}, and evaluating dialogue quality through MBTI-based categorization \citep{cheng-etal-2025-exploring} or other broad dimensions such as warmth and neuroticism \citep{choi2024examining}. However, these approaches rely on coarse-grained, personality-centric representations.

Other works have investigated persona-induced behavioral changes in specific settings, including psychological portrayal~\cite{huang2024on}, personality consistency in multi-agents \citep{frisch2024llm}, strategy shifts in emotional support dialogues \citep{wu-etal-2025-personas}, and sentiment polarity effects \citep{jun-lee-2025-exploring}. While informative, these works remain task-specific and do not identify which attributes mainly drive role-playing performance. Our schema supports field-level diagnosis of performance-driving attributes.


%
\section{Constructing a Controlled Character Profile Dataset}
\label{sec:dataset}
We build a standardized profile dataset isolating three axes---\textbf{Familiarity}, \textbf{Structure}, and \textbf{Disposition}---for controlled RPA analysis.



\subsection{Character Profile Schema}
\label{sec:profile_schema}
\begin{figure}[h]
    \centering
    
    \includegraphics[width=0.95\linewidth]{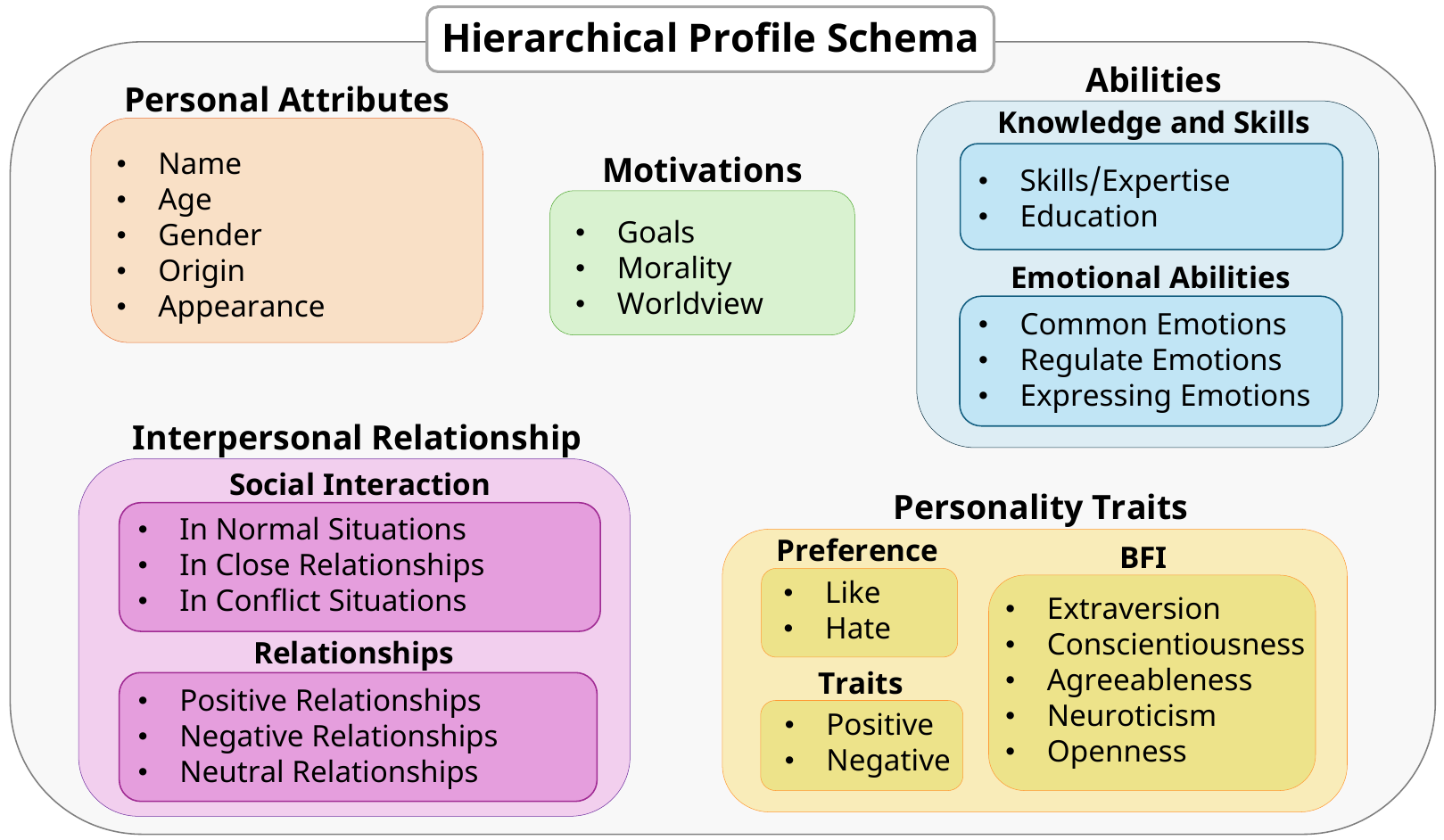}
    \vspace{-1mm}
    \caption{Hierarchical character profile schema: 28 leaf fields organized into 5 top-level dimensions.}
    \label{fig:schema_figure}
    \vspace{-4mm}
\end{figure}

We design a comprehensive character schema combining psychological theories, NLP evaluation frameworks, and practical RPA formats. Cognitive-affective theory~\cite{mischel1995cognitive} and the integrative personality framework~\cite{mcadams2006new} motivate high-level dimensions (traits, motivations, abilities, social behavior), while prior psychological evaluation of LLM personas~\cite{huang2024on} informs the overall organization. We inspect 50 popular user-crafted profiles from Risu AI\footnote{https://github.com/kwaroran/Risuai} and manually select dialogue-expressible fields mappable to these dimensions. As illustrated in Figure~\ref{fig:schema_figure}, this hierarchical schema consists of 5 top-level dimensions: \texttt{Personal Attributes}, \texttt{Personality Traits}, \texttt{Interpersonal Relationships}, \texttt{Motivations}, and \texttt{Abilities}. These expand into 15 mid-level dimensions and 28 leaf fields. See Appendix~\ref{app:Dataset_Schema} for the complete description.

\subsection{Dataset Construction}
\label{sec:data_construction}
We first construct character profiles along the \textbf{Familiarity} axis, then classify each profile based on the \textbf{Disposition} axis. Finally, we parallelize \textbf{Structure} variants from the resulting profiles.

\paragraph{Constructing \textit{Known} Character Set}
We use character pages from several famous works on Fandom\footnote{https://www.fandom.com/} as metadata, filtering noisy text (e.g., image hashes) via rule-based methods~\cite{NEURIPS2024_370df50c}. Following prior work showing that one-shot profiling via summarization is effective~\cite{yuan-etal-2024-evaluating}, we prompt Claude-4.5-Sonnet~\cite{anthropic_claude_sonnet_45_2025} to summarize this metadata into our template. Specifically, we select 3 to 4 major characters from 34 renowned works (movies, dramas, and animations), yielding 109 characters. Appendix~\ref{app:known_character} provides implementation details.

\paragraph{Constructing \textit{Unknown} Character Set}
We define \textit{Unknown} characters as newly synthesized, non-canonical personas that are not derived from any pre-existing character source. 
To align with the \textit{Known} construction, we synthesize \textit{Unknown} profiles via summarization rather than random field generation. We first define 11 variables — 9 demographic attributes, a personality-trait seed, and scenario genres with multiple candidates each, randomly sample combinations, and use Claude-4.5-Sonnet as an LLM-as-Judge to retain only coherent personas (score $\geq$ 8/10). We group these skeletons into sets of three and prompt GPT-oss-120B~\cite{agarwal2025gpt} to generate a story featuring them, then derive per-persona episodes as metadata. Finally, Claude-4.5-Sonnet summarizes each metadata into our schema template. In total, we generate 34 stories, produce 3 persona-specific episodes per story, and extract one character from each episode, yielding 102 \textit{Unknown} characters. Details are in Appendix~\ref{app:unknown_character}.

\paragraph{Classifying \textit{Moral}-\textit{Immoral} Characters}
To determine dispositional alignment, we utilize GPT-4o~\cite{hurst2024gpt} to assign a disposition score to each profile on a 10-point Likert scale. Characters scoring $\le$ 5 are classified as \textit{Immoral}, and those $\ge$ 6 as \textit{Moral}. 
We instruct the model to judge solely on the moral values and intrinsic nature described within the profile, disregarding canonical roles. 
This classification splits the 211 base characters nearly evenly into 106 \textit{Moral} and 105 \textit{Immoral} personas. Furthermore, we extend this evaluation beyond the holistic profile level to individual schema fields, constructing field-specific \textit{Moral} and \textit{Immoral} character sets for Section~\ref{sec:rq1_discussion}. See Appendix~\ref{app:moral_immoral_details} for details and human validation.

\paragraph{Parallelizing \textit{Structured}-\textit{Unstructured} Characters}
The aforementioned construction process yields a comprehensive set of \textit{Structured} profiles organized by our hierarchical schema. To obtain their \textit{Unstructured} counterparts, we convert each structured profile into a free-form narrative description while rigorously preserving the original semantic content. Specifically, we prompt the Claude-4.5-haiku model to rewrite the content of each structured profile into a continuous text format, entirely unconstrained by the original schema framework. We provide full prompt in Appendix~\ref{app:unstructured_details}.

\subsection{Profile Complexity Audit}
\begin{table}[!htbp]
\centering
\small
{\renewcommand{\arraystretch}{1.15}
\setlength{\tabcolsep}{3pt}
\begin{tabular}{l|cccc}
\hline
\textbf{Type} & \textbf{\# Characters} & \textbf{\# Words} & \textbf{\# Facts} & \textbf{Sim.} \\
\hline
Known    & 109 & 622.0 & 159.7 & 0.76 \\
Unknown & 102 & 551.3 & 134.8 & 0.75 \\
\midrule
Moral  & 106 & 609.2 & 154.9 & 0.75 \\
Immoral  & 105 & 570.3 & 142.1 & 0.76 \\
\midrule
\midrule
Structured & 211 & 589.8 & 148.5 & 0.76 \\
Unstructured  & 211 & 599.7 & 148.5 & 0.76 \\
\hline
\end{tabular}
}
\vspace{-1mm}
\caption{Dataset statistics broken down by profile type.}
\vspace{-2mm}
\label{tab:dataset_stats}
\end{table}

\label{sec:data_statistics}
To further characterize profile complexity across character types, we decompose each profile into atomic facts using Claude-Haiku-4.5 and report both the average number of extracted facts and the average pairwise embedding similarity among them. As shown in Table~\ref{tab:dataset_stats}, profiles contain approximately 550--620 words and 135--160 atomic facts on average. \textit{Known} profiles are slightly longer and more fact-dense than \textit{Unknown} profiles, and \textit{Moral} profiles show the same mild tendency compared to \textit{Immoral} profiles. 
Nevertheless, the differences are modest, and the average semantic similarity remains nearly identical across all profile types, ensuring that each axis is compared under matched profile complexity. Further dataset validations are in Appendix~\ref{app:data_validation}.

\section{Which Character Profile Axes Drive RPA Performance?}
\label{sec:rq1}
\begin{table*}
    \centering
    \small
    \begin{tabular}{@{}l|cccc|cccc|cccc}
    \toprule
        & \multicolumn{4}{c|}{\textbf{Familiarity}} & \multicolumn{4}{c|}{\textbf{Structure}}& \multicolumn{4}{c}{\textbf{Disposition}} \\
         & \textit{Known} & \textit{Unknown} & \textbf{$\Delta$} & \textbf{\textit{p}}  & \textit{Str.} & \textit{Unstr.} & \textbf{$\Delta$} & \textbf{\textit{p}} & \textit{Moral} & \textit{Immoral} & \textbf{$\Delta$} & \textbf{\textit{p}} \\
         \midrule
         \rowcolor{gray!30}
         \multicolumn{13}{c}{\textbf{PersonaGym (Single-turn interview)}} \\
    \midrule
         \textbf{Qwen3-8B} &

         4.50& 4.43 & -0.07 & *&
         4.46 & 4.47 & 0.01 & - &
         4.64&4.29& \textbf{-0.35}&***\\

         \textbf{Qwen3-235B} &

         4.66&4.62& -0.04 & -&
         4.64 & 4.61 & -0.03 & - &
         4.8 &4.47 & \textbf{-0.33}&***\\

         \textbf{Mistral-Small} &
         4.58&4.53 & -0.05 & * &
         4.55 & 4.54 & -0.01 & - &
         4.71&4.39 & \textbf{-0.32}&***\\

         \textbf{Olmo3.1-32B} &

         4.68 & 4.68 & 0.00 & - &
         4.68 & 4.62 & -0.07 & - &
         4.72 & 4.51 & \textbf{-0.20} & ** \\

         \textbf{DeepSeek-v3.2} &

         4.53 & 4.54 & 0.01 & -&
         4.54 & 4.51 & -0.03 & -&
         4.68 & 4.39 &\textbf{-0.29} & ***\\

         \midrule
    \rowcolor{gray!30}
    \multicolumn{13}{c}{\textbf{CoSER (Multi-turn interaction)}} \\
    \midrule
         \textbf{Qwen3-8B} &

         20.67 & 23.44 & 2.77 & - &
         22.06 & 22.37 & 0.31 & - &
         24.88 & 18.88 & \textbf{-6.00} & *** \\

         \textbf{Qwen3-235B} &

         33.30 & 35.00 & 1.70 & - &
         34.15 & 31.66 & -2.49 & ** &
         38.06 & 29.66 & \textbf{-8.40} & *** \\

         \textbf{Mistral-Small} &

         31.30 & 33.30 & 2.00 & - &
         32.30 & 33.84 & 1.54 & * &
         35.04 & 29.15 & \textbf{-5.89} & *** \\

         \textbf{Olmo3.1-32B} &

         22.77 & 24.78 & 2.01 & - &
         23.55 & 24.38 & 0.83 & - &
         26.81 & 20.28 & \textbf{-6.53} & *** \\

         \textbf{DeepSeek-v3.2} &

         38.10 & 42.63 & 4.53 & * &
         40.36 & 39.44 & -0.92 & - &
         44.82 & 35.60 & \textbf{-9.22} & *** \\
         \bottomrule
    \end{tabular}
    \vspace{-2mm}
    \caption{RPA performance across three axes (Familiarity, Structure,  Disposition) on PersonaGym and CoSER. ∆ denotes the difference between conditions; significance levels: * p<0.05, ** p<0.01, *** p<0.001.}
    \label{tab:rq1main}
\vspace{-3.5mm}
\end{table*}

Using the dataset from Section~\ref{sec:dataset}, we compare RPA performance across three axes: \textbf{Familiarity} (\textit{Known} vs.\ \textit{Unknown}), \textbf{Structure} (\textit{Structured} vs.\ \textit{Unstructured}), and \textbf{Disposition} (\textit{Moral} vs.\ \textit{Immoral}) under controlled, identical conditions.
\subsection{Experiment Setup}
\label{sec:rq1_setup}
\paragraph{Benchmarks} 
To evaluate the role-playing performance of RPAs, we adopt two benchmarks: PersonaGym~\cite{samuel2024personagym} for single-turn interviewing, and CoSER~\cite{wang2025coser} for multi-turn interactions. PersonaGym evaluates response quality via dynamic, persona-tailored interviews using five rubrics: \textit{Persona Consistency}, \textit{Linguistic Habits}, \textit{Expected Action}, \textit{Action Justification}, and \textit{Toxicity Control}. The averaged \textit{Persona Score} serves as our default metric on PersonaGym.

Next, CoSER assesses multi-party narratives expanded from seed scenarios using three rubrics: \textit{Anthropomorphism}, \textit{Character Fidelity}, and \textit{Storyline Quality}. The averaged \textit{Persona Score} serves as our default metric on CoSER. Since our dataset does not contain scenario information, we use gemini-2.5-pro~\cite{comanici2025gemini} to generate scenarios specifically tailored to the participating characters to serve as narrative seeds. To reduce scenario-specific bias, we evaluate each work with three distinct seed scenarios; each session uses three characters from the same material and runs for 18 turns. For the \textbf{Disposition} axis, CoSER sessions are labeled as \textit{Moral} when they include 0--1 immoral characters and \textit{Immoral} when they include 2--3. We provide further details in Appendix~\ref{app:RQ1_ExpSetup}.

\paragraph{Backbone LLMs}
We employ five recent open-source LLMs as our backbone role-playing models: Qwen3-8B, Qwen3-235B-A22B-Instruct~\cite{yang2025qwen3}, Mistral-Small-3.2-24B~\cite{liu2026ministral}, OLMo-3.1-32B-Instruct~\cite{olmo2025olmo}, and Deepseek-v3.2~\cite{liu2025deepseek}. For quantitative evaluation on both benchmarks, we utilize GPT-4o~\cite{hurst2024gpt} as the Judge.

\subsection{Main Results}
\label{sec:rq1_result}

Table~\ref{tab:rq1main} presents RPA performance across all three axes. The results reveal a clear asymmetry: \textit{Familiarity} and \textit{Structure} show negligible impact, while \textit{Disposition} produces large and consistent performance degradation across all models. We further validate our main findings by human evaluation, which exhibits an identical trend (see Appendix~\ref{app:rq1_human} for details). We also provide per-metric breakdowns in Appendix~\ref{app:per_axis}.

\paragraph{Familiarity: Task-dependent but not a stable driver.}
The \textbf{Familiarity} axis exhibits a benchmark-dependent pattern rather than a stable advantage in either direction. On PersonaGym, \textit{Known} characters obtain slightly higher scores for some models, but the effect sizes remain very small ($|\Delta| \leq 0.07$). On CoSER, \textit{Unknown} characters show a higher-score tendency across models, but only DeepSeek-v3.2 reaches statistical significance ($p<0.05$). We interpret this reversal as an interaction between canonical priors and contextual grounding: prior knowledge may help in single-turn interviews, while accumulated profile and scenario context in multi-turn narratives can dilute or compete with such priors. Appendix~\ref{app:known_unknown} further supports this interpretation through a turn-wise analysis of the Familiarity axis.

\paragraph{Structure: Marginal and inconsistent.}
Profile \textbf{Structure} shows virtually no impact on performance. On PersonaGym, the gap between \textit{Structured} and \textit{Unstructured} profiles is at most 0.03 and never reaches statistical significance. On CoSER, while Qwen3-235B and Mistral-Small show marginal significance, the effect sizes remain small ($\Delta \leq 2.49$), and the direction is inconsistent. Structured formatting does not provide a meaningful advantage over free-form narratives. We emphasize that these Familiarity and Structure null results hold under controlled conditions in which every profile is rich, schema-based, and semantically matched. They characterize the marginal effect of each axis once such profile information is supplied, and do not imply that familiarity or format is irrelevant under sparser or lower-quality profiles.

\paragraph{Disposition: Large and consistent degradation.}
In stark contrast, \textbf{Disposition} axis produces the largest and most consistent performance gap across all conditions. On PersonaGym, \textit{Immoral} characters score significantly lower than \textit{Moral} ones across every model ($\Delta$: 0.20--0.35, all $p < 0.01$). On CoSER, the gap widens further ($\Delta$: 5.89--9.22, all $p < 0.001$), with DeepSeek-v3.2 exhibiting the largest drop of 9.22 points. Unlike the other two axes, disposition effect is uniformly significant, large in magnitude, and consistent in direction across both benchmarks and all models.

\subsection{Analyses \& Discussion}
\label{sec:rq1_discussion}

\paragraph{Counterfactual rewrite validation.}
\begin{table}[t]
\centering
\small
\resizebox{0.9\columnwidth}{!}{
\begin{tabular}{l|cc||cc}
\toprule
\multirow{2}{*}{\textbf{Model}} 
& \multicolumn{2}{c||}{\textbf{Familiarity}} 
& \multicolumn{2}{c}{\textbf{Disposition}} \\
& \textbf{\textit{K}} & \textbf{\textit{UK}} & \textbf{\textit{M}} & \textbf{\textit{IM}} \\
\midrule

\rowcolor{gray!30}
\multicolumn{5}{c}{\textbf{PersonaGym}} \\
\midrule

\multirow{1}{*}{Qwen3-235B} 
& -0.17
& -0.18 
& \textbf{-0.82} 
& \textbf{+0.26} \\
\midrule

\multirow{1}{*}{DeepSeek-v3.2} 
& -0.16 
& -0.11 
& \textbf{-0.62} 
& \textbf{+0.34} \\

\midrule
\rowcolor{gray!30}
\multicolumn{5}{c}{\textbf{CoSER}} \\
\midrule

\multirow{1}{*}{Qwen3-235B} 
& -1.02 
& +0.81 
& \textbf{-22.17} 
& \textbf{+3.47} \\
\midrule

\multirow{1}{*}{DeepSeek-v3.2} 
& -0.55 
& -0.21 
& \textbf{-19.89} 
& \textbf{+1.79} \\
\bottomrule
\end{tabular}
}
\vspace{-2mm}
\caption{Counterfactual rewrite results. K and UK denote \textit{Known} and \textit{Unknown}, respectively, while M and IM denote \textit{Moral} and \textit{Immoral}.}
\label{tab:counterfactual_main}
\vspace{-4mm}
\end{table}

To examine whether the observed gaps follow the intended profile axes rather than fixed character identities, we conduct counterfactual rewrite experiments on shared base profiles. For the \textbf{Familiarity} axis, we swap the \texttt{Personal Attributes} field between paired \textit{Known} and \textit{Unknown} profiles; for the \textbf{Disposition} axis, we rewrite the remaining fields toward the opposite moral polarity. Table~\ref{tab:counterfactual_main} reports the score change from the original to the counterfactual profile. The Familiarity swap yields negligible changes, within $\pm0.2$ on PersonaGym and $\pm1.1$ on CoSER. In contrast, the Disposition rewrite produces large directional shifts: \textit{Moral}$\rightarrow$\textit{Immoral} sharply degrades performance, up to $-0.82$/$-22.17$ on PersonaGym/CoSER, while \textit{Immoral}$\rightarrow$\textit{Moral} improves it, up to $+0.34$/$+3.47$. Together, these results support our main finding that \textbf{Familiarity} is not a major driver, while the \textbf{Disposition} gap follows dispositional content rather than character identity. To further isolate value-laden polarity, we extend this analysis to single-field edits. Holding all other content fixed, we rewrite only one value-laden field—the \texttt{Motivations} dimension or its \texttt{Morality}/\texttt{Goal} leaves—to the opposite polarity. A single-field flip reproduces the directional pattern on both benchmarks and recovers roughly half of the full-rewrite effect, preserving the same asymmetry in which degradation exceeds recovery. Because every co-occurring content factor is held constant by design, this attributes a substantial portion of the gap to value-laden polarity itself. Detailed procedure and results are provided in Appendix~\ref{app:rewrite_prompt}.

\paragraph{Alignment-stage ablation.}
\begin{table}[t]
\centering
\small
\resizebox{0.8\columnwidth}{!}{
\begin{tabular}{l|cccc}
\toprule
\textbf{Rubrics}
& \textbf{\textit{M}} 
& \textbf{\textit{IM}}
& \textbf{$\Delta$}
& \textbf{\textit{p}} \\
\midrule

\rowcolor{gray!20}
\multicolumn{5}{c}{\textbf{Instruct-SFT}} \\
\midrule

\textit{Anthro.} 
& 10.78 & 6.02 & -4.76 & * \\
\textit{Char. Fid.} 
& 15.88 & 13.44 & -2.44 & - \\
\textit{Story Qual.} 
& 32.45 & 26.66 & -5.79 & * \\
\midrule
\textit{Avg.}
& 19.70 & 15.37 & -4.53 & * \\
\midrule

\rowcolor{red!15}
\multicolumn{5}{c}{\textbf{$+$ DPO}} \\
\midrule

\textit{Anthro.} 
& 13.53 & 6.32 & \textbf{-7.21} & *** \\
\textit{Char. Fid.} 
& 16.67 & 7.01 & \textbf{-9.66} & *** \\
\textit{Story Qual.} 
& 46.08 & 39.95 & -6.13 & ** \\
\midrule
\textit{Avg.}
& 25.42 & 17.79 & \textbf{-7.63} & *** \\

\bottomrule
\end{tabular}
}
\vspace{-2mm}
\caption{Alignment-stage ablation on CoSER. \textit{M}, \textit{IM} denote \textit{Moral}, \textit{Immoral}; \textit{Anthro.}, \textit{Char. Fid.}, and \textit{Story Qual.} denote \textit{Anthropomorphism}, \textit{Character Fidelity}, and \textit{Storyline Quality}. Significance levels: * $p<0.05$, ** $p<0.01$, *** $p<0.001$.}
\label{tab:dpo_ablation}
\vspace{-5mm}
\end{table}

Because the \textbf{Disposition} axis is conceptually tied to post-SFT alignment, we further examine whether alignment amplifies the Moral--Immoral asymmetry. To this end, we compare two checkpoints from the same model family: OLMo-3.1-32B-Instruct-SFT before preference alignment and OLMo-3.1-32B-Instruct-DPO after DPO alignment~\cite{rafailov2023direct}. As shown in Table~\ref{tab:dpo_ablation}, DPO improves the average CoSER scores for both \textit{Moral} and \textit{Immoral} characters, but the improvement is larger for \textit{Moral} characters, widening the average Moral--Immoral gap from $-4.53$ to $-7.63$. This widening is most pronounced in \textit{Character Fidelity}: the gap is small and not significant under Instruct-SFT ($\Delta=-2.44$), but becomes large and significant after DPO ($\Delta=-9.66$). Notably, while DPO slightly improves \textit{Character Fidelity} for \textit{Moral} characters, it substantially decreases it for \textit{Immoral} characters. These results suggest that DPO-style preference alignment, which favors human-preferred helpful or benign responses, can selectively weaken fidelity to immoral profile content. We further evaluate a subsequent RLVR-aligned checkpoint~\cite{lambert2025tulu} and find that it partially improves overall role-playing scores, but does not close the DPO-amplified Moral--Immoral gap; see Appendix~\ref{app:rlvr_ablation}.
We note that the residual gaps under Instruct-SFT and after FACD need not stem from alignment alone. Immoral profiles also tend to encode more aversive content (e.g., conflict, manipulation, harm), which may independently increase modeling difficulty \citep{yi2026too}.

\paragraph{Field-level localization of the Disposition bottleneck.}
\begin{figure}[t]
    \centering
    \includegraphics[width=1\columnwidth]{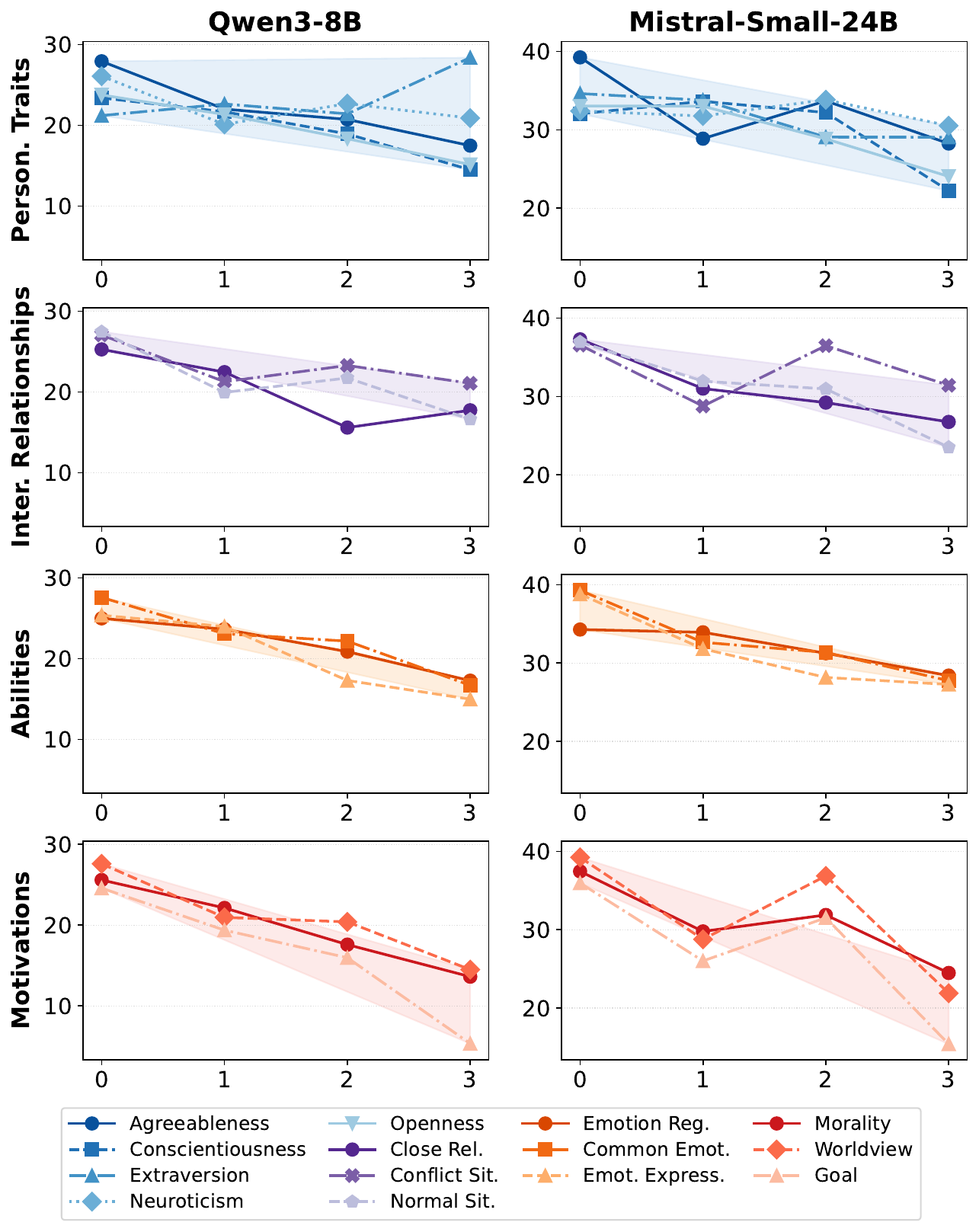}
    \vspace{-3mm}
    \caption{Field-level localization of the Disposition bottleneck on CoSER. X-axis denotes the number of \textit{Immoral} characters included in each sample, and the Y-axis denotes the \textit{CoSER Score}.}
    \label{fig:rq2_Coser}
    \vspace{-5mm}
\end{figure}

Building on the finding that the \textbf{Disposition} axis drives the largest performance disparities, we further examine whether this degradation is holistic or localized to specific profile dimensions. Using the field-categorized character set (Section~\ref{sec:data_construction}), we track CoSER performance as the number of \textit{Immoral} characters increases.

As shown in Figure~\ref{fig:rq2_Coser}, the degradation is not uniform across fields. \texttt{Personality Traits} remain relatively stable, suggesting that models can still reproduce surface-level dispositional signals such as extraversion or agreeableness. In contrast, fields in \texttt{Motivations} exhibit the steepest and most consistent decline, especially in value-laden fields such as \textit{Morality} and \textit{Goal}. Fields in \texttt{Abilities} and \texttt{Interpersonal Relationships} show intermediate degradation. These results indicate that the primary bottleneck lies not in adopting basic character traits, but in faithfully representing specific value-laden profile fields under \textit{Immoral} settings. This motivates the targeted mitigation strategy introduced in Section~\ref{sec:rq2}. Full localization results across all five models are provided in Appendix~\ref{app:localization_full}.

\section{Mitigating the Alignment Bottleneck}
\label{sec:rq2}

Our analysis shows that this bottleneck is concentrated in specific value-laden fields, including \texttt{Motivations} fields such as \textit{Goal} and \textit{Morality}, as well as disposition-sensitive fields in other profile dimensions. Now, our objective is to improve the fidelity of \textit{Immoral} characters without compromising the baseline performance of \textit{Moral} ones. To this end, we introduce \textit{Field-Aware Contrastive Decoding} (FACD), a training-free strategy that contrasts the full profile against a sanitized profile in which immoral or disposition-sensitive fields are removed, thereby selectively amplifying the suppressed profile signals at inference time. We further validate our results by human evaluation, which exhibits an identical trend (see Appendix~\ref{app:rq2_human} for details).

\subsection{Preliminary: Contrastive Decoding}
\label{sec:preliminary_cd}
Modern alignment procedures instill strong helpful-assistant priors into LLMs~\citep{NEURIPS2022_b1efde53, huang2025safety}, which persistently resist conflicting behavioral instructions delivered through system prompts~\citep{sharma2024towards}. \textit{Contrastive Decoding}~\citep{li-etal-2023-contrastive} offers a principled remedy by reformulating generation as a contrast between two output distributions, thereby isolating and amplifying a desired behavioral signal at decoding time without any parameter updates.

Formally, consider an autoregressive language model $M_\theta$ that produces next-token logits $z_t = \log p_\theta(x_t \mid \mathbf{x}_{<t})$ given a prefix $\mathbf{x}_{<t}$.
\cite{dong2026steer} adapts the contrastive principle to system-prompt adherence by running two parallel forward passes with different system prompts: a \emph{positive prompt} $s_{\mathrm{pos}}$ encoding the desired persona, and a \emph{negative prompt} $s_{\mathrm{neg}}$ representing the behavior we wish to suppress.
Let $z_t^{\mathrm{pos}}$ and $z_t^{\mathrm{neg}}$ denote the logits produced under each prompt, respectively. The steered logits are then defined as:
\begin{equation}
\label{eq:cd}
    \tilde{z}_t = z_t^{\mathrm{pos}} + \alpha\, (z_t^{\mathrm{pos}}-z_t^{\mathrm{neg}}),
\end{equation}

where $\alpha$ controls the amplification strength. At $\alpha{=}0$, generation reduces to standard decoding under $s_{\mathrm{pos}}$. As $\alpha$ increases, the contrast term $z_t^{\mathrm{pos}}-z_t^{\mathrm{neg}}$ is amplified, suppressing tokens favored by the sanitized baseline ($s_{\mathrm{neg}}$) while boosting tokens specific to the desired persona ($s_{\mathrm{pos}}$).

\subsection{Field-Aware Contrastive Decoding}
\label{sec:facd}

\paragraph{Motivation.} 
While standard contrastive decoding uniformly amplifies the entire persona signal, the bottleneck identified in Section~\ref{sec:rq1} requires a more targeted intervention. When a profile prescribes adversarial traits, the alignment prior actively suppresses the specific tokens needed to portray them. Crucially, this suppression is not uniform across the profile: fields like \texttt{Goal} suffer steep declines, whereas polarity-insensitive fields like \texttt{Extraversion} remain stable (Figure~\ref{fig:rq2_Coser}). Therefore, we propose a \textit{field-aware} variant that directs contrastive amplification exclusively toward the immoral signals that alignment suppresses.

\paragraph{Method.}
Let $\mathcal{F}$ denote the set of all fields in a character profile, and let $s(\mathcal{F}')$ denote the full role-playing prompt whose character profile is restricted to a field subset $\mathcal{F}' \subseteq \mathcal{F}$. The positive prompt $s_{\mathrm{pos}} = s(\mathcal{F})$ uses the complete profile.

FACD constructs the negative prompt as a sanitized profile baseline. Starting from the full profile $\mathcal{F}$, we remove all non-Personal-Attributes fields classified as immoral by MoralBERT~\citep{preniqi2024moralbert}. Let $\mathcal{F}_{\mathrm{PA}}$ denote the Personal Attributes fields and
$\mathcal{F}_{\mathrm{immoral}}=\{f \in \mathcal{F}\setminus\mathcal{F}_{\mathrm{PA}}: M(f)=\mathrm{immoral}\}$,
where $M$ is MoralBERT. The retained fields are $\mathcal{F}_{\mathrm{base}} = \mathcal{F} \setminus \mathcal{F}_{\mathrm{immoral}}$.

If fewer than 6 non-PA fields remain after filtering, we add a fallback set $\mathcal{F}_{\mathrm{pad}}$ drawn from disposition-robust fields identified in Figure~\ref{fig:rq2_Coser}, such as \texttt{Extraversion}, \texttt{Neuroticism}, and \texttt{Openness}. The final negative prompt is $s_{\mathrm{neg}} = s\bigl(\mathcal{F}_{\mathrm{base}} \cup \mathcal{F}_{\mathrm{pad}}\bigr)$. Thus, $s_{\mathrm{neg}}$ is a sanitized version of the original profile produced after both immoral-field filtering and disposition-robust padding. As a result of this full construction pipeline, the final $s_{\mathrm{neg}}$ contains no field that is both disposition-sensitive and immoral. The steered logits follow Eq.~\ref{eq:cd}; the contrast $z_t^{\mathrm{pos}}-z_t^{\mathrm{neg}}$ therefore amplifies the profile signals contributed by the omitted immoral fields, especially when they occur in disposition-sensitive parts of the profile. We set $\alpha{=}1$ as a moderate value that empirically balances fidelity gains and narrative coherence. We provide more details about FACD in Appendix~\ref{app:facd_details}.

\subsection{Mitigation Results}
\label{sec:rq2_results}
\begin{figure}[!htbp]
\centering

\begin{subfigure}{\linewidth}
    \centering
    \scriptsize
    \setlength{\tabcolsep}{2.8pt}
    \renewcommand{\arraystretch}{1.03}

    \resizebox{\linewidth}{!}{%
    \begin{tabular}{l|ccc|c||ccc|c}
    \toprule
    \multirow{2}{*}{\textbf{Dim.}} & \multicolumn{4}{c||}{\textbf{Qwen3-8B}} & \multicolumn{4}{c}{\textbf{Mistral-Small}} \\
    & M & IM & $\Delta$ & \textit{Avg.} & M & IM & $\Delta$ & \textit{Avg.} \\
    \midrule
    \rowcolor{gray!20}
    \multicolumn{9}{c}{\textbf{Default}} \\
    \midrule
    \textit{Anthro.}
    & 14.53 & 8.55 & {\scriptsize-5.98} & 11.72
    & 20.27 & 17.85 & {\scriptsize-2.42} & 19.13 \\
    \textit{Char. Fid.}
    & 11.39 & 6.39 & {\scriptsize-5.00} & 9.04
    & 27.12 & 15.78 & {\scriptsize-11.34} & 21.78 \\
    \textit{Story Qual.}
    & 48.72 & 41.71 & {\scriptsize-7.01} & 45.42
    & 57.74 & 53.82 & {\scriptsize-3.92} & 55.90 \\
    \midrule
    \textit{Avg.}
    & 24.88 & 18.88 & {\scriptsize-6.00} & 22.06
    & \textbf{35.04} & 29.15 & {\scriptsize-5.89} & 32.27 \\
    \midrule
    \rowcolor{pink!30}
    \multicolumn{9}{c}{\textbf{Field-aware Contrastive Decoding (FACD)}} \\
    \midrule
    \textit{Anthro.}
    & \textbf{18.24} & \textbf{11.39} & {\scriptsize-6.85} & \textbf{15.02}
    & \textbf{21.01} & \textbf{18.37} & {\scriptsize-2.64} & \textbf{19.77} \\
    \textit{Char. Fid.}
    & \textbf{13.91} & \textbf{14.38} & {\scriptsize\textbf{+0.47}} & \textbf{14.13}
    & 26.22 & \textbf{24.38} & {\scriptsize\textbf{-1.84}} & \textbf{25.35} \\
    \textit{Story Qual.}
    & \textbf{49.06} & \textbf{41.95} & {\scriptsize-7.11} & \textbf{45.71}
    & 55.71 & 53.52 & {\scriptsize\textbf{-2.19}} & 54.68 \\
    \midrule
    \textit{Avg.}
    & \textbf{27.07} & \textbf{22.57} & {\scriptsize\textbf{-4.50}} & \textbf{24.95}
    & 34.31 & \textbf{32.09} & {\scriptsize\textbf{-2.22}} & \textbf{33.27} \\
    \bottomrule
    \end{tabular}%
    }
\end{subfigure}

\vspace{1mm}

\begin{subfigure}{\linewidth}
    \centering
    \includegraphics[width=\linewidth]{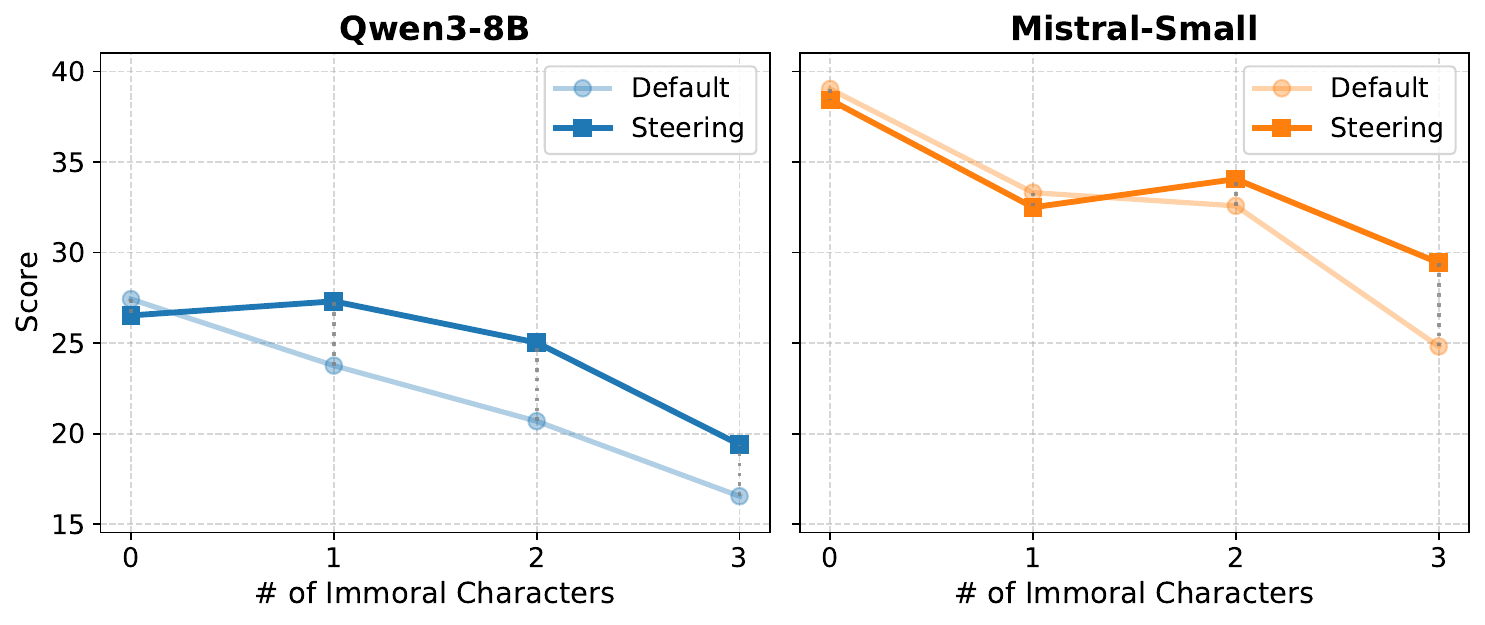}
\end{subfigure}

\vspace{-2mm}
\caption{Comparison of steering performance and mitigation results across models. Y-axis of the lower figure indicates the CoSER score.}
\label{fig:mitigation}
\vspace{-4mm}
\end{figure}

\paragraph{FACD narrows the \textit{Moral}--\textit{Immoral} gap.}
As shown in the top table of Figure~\ref{fig:mitigation}, the default setting yields a substantial performance gap between \textit{Moral} (M) and \textit{Immoral} (IM): $-6.00$ for Qwen3-8B and $-5.89$ for Mistral-Small on average. Applying FACD reduces these gaps to $-4.50$ and $-2.22$, respectively. Importantly, this improvement does not come from sacrificing \textit{Moral} performance for \textit{Immoral} recovery: the \textit{Moral} average increases for Qwen3-8B ($24.88 \rightarrow 27.07$) and remains comparable for Mistral-Small ($35.04 \rightarrow 34.31$). These results indicate that FACD mitigates the dispositional asymmetry while preserving overall role-playing quality, and can even benefit \textit{Moral} settings when value-laden profile fields are under-expressed under default decoding.

\paragraph{Character Fidelity shows targeted recovery.}
The largest recovery appears in \textit{Character Fidelity}, the metric most directly tied to adherence to the given profile. This matches our diagnosis that the bottleneck primarily concerns the under-expression of immoral content in the disposition-sensitive profile field. On Qwen3-8B, the IM Character Fidelity score more than doubles from $6.39$ to $14.38$, closing the gap from $-5.00$ to $+0.47$. On Mistral-Small, it rises from $15.78$ to $24.38$, reducing the gap from $-11.34$ to $-1.84$. Meanwhile, \textit{Anthropomorphism} and \textit{Storyline Quality} largely preserved on both models. Thus, FACD does not merely make generations more intense or broadly persona-like; it primarily recovers profile-grounded traits that are otherwise under-expressed.

\paragraph{Effective under stronger dispositional pressure.}
The lower panel of Figure~\ref{fig:mitigation} shows that FACD is beneficial as the number of \textit{Immoral} characters increases. Although the downward trend is not fully eliminated, FACD consistently lifts performance above the default baseline, with the largest gains appearing when all three characters are \textit{Immoral}: Qwen3-8B improves from roughly 16 to 20, and Mistral-Small from 25 to 29. This indicates that FACD is most valuable in challenging multi-character settings, where the need to jointly instantiate multiple immoral personas makes the dispositional bottleneck more severe.

\paragraph{Judge validity.} These results also address a concern that the judge might reward safe or prosocial behavior rather than profile adherence. If so, FACD—which amplifies immoral profile signals—would lower scores; instead it raises Immoral Character Fidelity (6.39 to 14.38 on Qwen3-8B), with gains concentrated in Character Fidelity rather than Anthropomorphism or Storyline Quality. Combined with the alignment-stage dependence of the gap in Table~\ref{tab:dpo_ablation}, this indicates the judge responds to profile compliance rather than likability or safety.

\subsection{Ablation Study}
\begin{table}[t]
\centering
\small
\setlength{\tabcolsep}{7pt}
\renewcommand{\arraystretch}{1.12}
\resizebox{\linewidth}{!}{%
\begin{tabular}{l|ccc|c}
\toprule
\textbf{Baselines} & \textbf{M} & \textbf{IM} & \textbf{$\Delta$} & \textbf{\textit{Avg.}} \\
\midrule
\textit{SICP} & 23.33 & 19.70  & -3.63 & 21.52 \\
\midrule
\textit{Personal Attributes} only & 22.02 & 18.40 & -3.62 & 20.21 \\
w/o. \textit{random fields} & 24.37 & 19.23  & -5.14 & 21.80 \\
\midrule
w/o. \textit{Personality Traits} & 23.27 & 18.81 & -4.46 & 21.04 \\
w/o. \textit{Interpersonal Relationships} & 23.18 & 20.53 & -2.65 & 21.86 \\
w/o. \textit{Motivations} & 25.64 & 20.42 & -5.22 & 23.03 \\
w/o. \textit{Abilities} & 24.61 & 19.60 & -5.01 & 22.11 \\
\midrule
w/o. \textit{Value-laden fields} & 24.75 & 21.11 & -3.64 & 22.93 \\
\midrule
\rowcolor{gray!15}
\textbf{\textit{FACD}} & \textbf{27.07} & \textbf{22.57} & -4.50 & \textbf{24.82} \\
\bottomrule
\end{tabular}%
}
\caption{Ablation study on the negative prompt construction for FACD. \textit{M} and \textit{IM} denote \textit{Moral} and \textit{Immoral}, respectively. $\Delta$ denotes the Moral--Immoral gap, and \textit{Avg.} denotes the mean of $M$ and $IM$.}
\vspace{-5mm}
\label{tab:facd_ablation}
\end{table}

We ablate the construction of the FACD negative prompt by retaining only \textit{Personal Attributes}, removing each top-level field group, or removing all disposition-sensitive fields regardless of Moral or Immoral polarity. As shown in Table~\ref{tab:facd_ablation}, the \textit{Personal Attributes only} baseline yields low scores for both groups, suggesting that an overly sparse negative prompt fails to provide a sufficiently character-grounded contrast. Beyond field removal, we also compare prompt-level and polarity-agnostic alternatives. Stay-In-Character Prompting (SICP), which reinforces disposition adherence purely through instructions, attains a smaller gap only because it degrades Moral performance rather than recovering Immoral fidelity, consistent with evidence that prompting alone cannot override alignment priors \citep{sharma2024towards}. Constructing the negative prompt from randomly selected fields (\emph{w/o.\ random fields}) yields limited gains, since without polarity awareness immoral signals are amplified or suppressed without a consistent criterion. Both reinforce the core design of FACD. An effective negative prompt must be simultaneously character-grounded and polarity-aware. 

Among the field-removal variants, excluding more disposition-sensitive dimensions such as \textit{Motivations} and \textit{Abilities} is more effective than excluding relatively surface-level \textit{Personality Traits}. Although some ablations produce a smaller Moral--Immoral gap, this mainly reflects degraded \textit{Moral} performance rather than improved \textit{Immoral} recovery, making such reductions less practical. Removing all value-laden fields also performs competitively, but still falls short of FACD. This supports the core design of FACD: an effective negative prompt should be both character-grounded and polarity-aware, preserving enough profile context while selectively omitting immoral/disposition-sensitive fields. This allows the contrastive term to amplify profile-specific signals from the full profile rather than generic assistant behavior.

\begin{table}[t]
\centering
\small
\setlength{\tabcolsep}{7pt}
\renewcommand{\arraystretch}{1.08}

\resizebox{\linewidth}{!}{%
\begin{tabular}{llccc}
\toprule
\textbf{Model} & \textbf{Type}
& \textbf{M (\%)}
& \textbf{IM (\%)}
& \cellcolor{gray!15}\textbf{\textit{Avg}. (\%)} \\
\midrule

\multirow{3}{*}{Qwen3-8B}
& Default              & 0.082 & 0.304 & \cellcolor{gray!15}0.193 \\
& FACD                 & 0.147 & 0.687 & \cellcolor{gray!15}0.417 \\
\cmidrule(lr){2-5}
& \textit{$\Delta$pp}  & 0.065 & 0.383 & \cellcolor{gray!15}0.224 \\

\midrule

\multirow{3}{*}{Mistral-Small}
& Default              & 0.000 & 0.221 & \cellcolor{gray!15}0.110 \\
& FACD                 & 0.146 & 0.495 & \cellcolor{gray!15}0.320 \\
\cmidrule(lr){2-5}
& \textit{$\Delta$pp}  & 0.146 & 0.274 & \cellcolor{gray!15}0.210 \\

\bottomrule
\end{tabular}%
}

\caption{
Utterance-level unsafe-content ratios (\%) under default decoding and FACD.
M and IM denote Moral and Immoral characters, respectively.
$\Delta$pp denotes the percentage-point difference between FACD and Default.
}
\label{tab:safety}
\end{table}

\subsection{Safety Analysis}

Because FACD amplifies disposition-sensitive signals, we verify that it does not introduce actionable harmful content, measuring fidelity and harmfulness with two independent instruments. GPT-4o for role-playing quality and Llama-Guard-4-12B\footnote{\url{https://huggingface.co/meta-llama/Llama-Guard-4-12B}} for safety. We split each simulated log into per-character utterances and judge each target utterance with its preceding turns as context, excluding flags triggered by non-content categories. As shown in Table~\ref{tab:safety}, FACD raises the flagged ratio only marginally: absolute rates remain below 0.7\% and all increases are under 0.4pp. This indicates that FACD amplifies profile-specified fictional signals rather than actionable harmful content; conversely, default decoding suppresses even content that is not actually unsafe. We discuss the resulting fidelity–safety trade-off in the Ethics Statement.

\section{Conclusion}
We present a diagnostic framework that disentangles role-playing fidelity along three axes—Familiarity, Structure, and Disposition—revealing that moral Disposition is the primary bottleneck. The resulting Moral–Immoral gap is substantially amplified by post-SFT alignment, which suppresses tokens needed for immoral portrayal.
To address this, we propose Field-Aware Contrastive Decoding, a training-free strategy that selectively amplifies suppressed traits, significantly narrowing the Moral–Immoral gap without fine-tuning.

\section*{Limitations}
While our study provides a foundational disentanglement of character profile axes, it has two limitations that offer avenues for future work. First, we adopted a binary classification of \textbf{Disposition}. Although this simplified categorization was essential for our diagnostic study to identify a substantial alignment-attributable component of the Moral–Immoral gap, it may not fully capture the complexity of "anti-heroes" or morally ambiguous personas found in nuanced narratives. Second, due to computational resource constraints inherent in running parallel forward passes for contrastive decoding, our proposed Field-Aware Contrastive Decoding (FACD) was evaluated on a subset of the frontier LLMs. Despite these constraints, the consistent performance gains observed in the evaluated models underscore FACD's potential as a robust, training-free mitigation strategy, providing a strong basis for future scaling to even larger model architectures. Finally, our safety assessment operates at the utterance level and therefore does not capture persuasive or normalizing effects that may accumulate across long, multi-turn interactions; evaluating such long-horizon harms remains an open problem.

\section*{Ethics Statement}
This paper compares role-playing behavior under two sources of character information: (i) Known character profiles derived from publicly accessible, community-curated Fandom pages, and (ii) Unknown character profiles generated by an LLM (gpt-oss). We provide clear attribution for third-party sources and document our generation procedure; we release only the minimum derived, structured metadata needed for research rather than reproducing copyrighted source text verbatim, and we expect downstream users to comply with applicable licenses/terms. Because both fan-curated metadata and model generation may contain or imply harmful content (e.g., violence, harassment, stereotyping), we apply filtering and normalization using a safety-aligned Claude model and conduct manual spot checks, while noting that residual risks may remain. Our evaluation instantiates these profiles within \textsc{CoSER}~\cite{wang2025coser} and \textsc{PersonaGym}~\cite{samuel2024personagym} pipelines and follows their ethical framing: persona/role-playing systems can be misused to generate targeted harmful content, reinforce stereotypes, or encourage anthropomorphization, and persona construction may raise intellectual-property and privacy concerns (especially if adapted to real individuals), which would require explicit consent and stronger protections. Finally, since we use LLMs for generation and/or automated assessment, we acknowledge potential judge/model biases and report configurations to support reproducibility, and we do not position these methods for safety-critical deployment. Our simulation logs are model–model interactions. A malicious user directly steering an immoral character could exploit it for harmful ends, a risk that grows once FACD strengthens immoral fidelity, so the method calls for ethically bounded use. Second, our utterance-level safety judging captures the harmfulness of individual turns but does not measure persuasive or normalizing effects that accumulate across long conversations, and evaluating such long-horizon harms remains an open problem. Third, FACD amplifies signals already specified in the character profile rather than generating arbitrary harmful content—its contrast is bounded by a sanitized version of the same profile—yet we acknowledge its dual-use potential. We therefore treat the fidelity–safety trade-off as an explicit deployment consideration.

\section*{Acknowledgement}
This work was supported by the Institute of Information \& Communications Technology Planning \& Evaluation (IITP) grant funded by the Korea government (MSIT) [RS-2021-II211341, Artificial Intelligence Graduate School Program (Chung-Ang University)]. This work was also supported by Korea Institute for Advancement of Technology(KIAT) grant funded by the Korea Government(MOTIE) (RS-2025-25458133) and the National Research Foundation of Korea(NRF) grantfunded by the Korea government(MSIT) (RS-2026-25494299).

\bibliography{acl_latex}

@article{wang2025opencharacter,
  title={Opencharacter: Training customizable role-playing llms with large-scale synthetic personas},
  author={Wang, Xiaoyang and Zhang, Hongming and Ge, Tao and Yu, Wenhao and Yu, Dian and Yu, Dong},
  journal={arXiv preprint arXiv:2501.15427},
  year={2025}
}

@inproceedings{zhou2025characterbench,
  title={CharacterBench: Benchmarking Character Customization of Large Language Models},
  author={Zhou, Jinfeng and Huang, Yongkang and Wen, Bosi and Bi, Guanqun and Chen, Yuxuan and Ke, Pei and Chen, Zhuang and Xiao, Xiyao and Peng, Libiao and Tang, Kuntian and others},
  booktitle={Proceedings of the AAAI Conference on Artificial Intelligence},
  volume={39},
  number={24},
  pages={26101--26110},
  year={2025}
}

@article{choi2024examining,
  title={Examining Identity Drift in Conversations of LLM Agents},
  author={Choi, Junhyuk and Hong, Yeseon and Kim, Minju and Kim, Bugeun},
  journal={arXiv preprint arXiv:2412.00804},
  year={2024}
}

@inproceedings{huang2024on,
title={On the Humanity of Conversational {AI}: Evaluating the Psychological Portrayal of {LLM}s},
author={Jen-tse Huang and Wenxuan Wang and Eric John Li and Man Ho LAM and Shujie Ren and Youliang Yuan and Wenxiang Jiao and Zhaopeng Tu and Michael Lyu},
booktitle={The Twelfth International Conference on Learning Representations},
year={2024},
url={https://openreview.net/forum?id=H3UayAQWoE}
}

@article{frisch2024llm,
  title={LLM agents in interaction: Measuring personality consistency and linguistic alignment in interacting populations of large language models},
  author={Frisch, Ivar and Giulianelli, Mario},
  journal={arXiv preprint arXiv:2402.02896},
  year={2024}
}

@inproceedings{cheng-etal-2025-exploring,
    title = "Exploring Personality-Aware Interactions in Salesperson Dialogue Agents",
    author = "Cheng, Sijia  and
      Chang, Wen Yu  and
      Chen, Yun-Nung",
    editor = "Torres, Maria Ines  and
      Matsuda, Yuki  and
      Callejas, Zoraida  and
      del Pozo, Arantza  and
      D'Haro, Luis Fernando",
    booktitle = "Proceedings of the 15th International Workshop on Spoken Dialogue Systems Technology",
    month = may,
    year = "2025",
    address = "Bilbao, Spain",
    publisher = "Association for Computational Linguistics",
    url = "https://aclanthology.org/2025.iwsds-1.6/",
    pages = "60--71",
    ISBN = "979-8-89176-248-0"
}

@inproceedings{jiang2024personallm,
  title={PersonaLLM: Investigating the ability of large language models to express personality traits},
  author={Jiang, Hang and Zhang, Xiajie and Cao, Xubo and Breazeal, Cynthia and Roy, Deb and Kabbara, Jad},
  booktitle={Findings of the association for computational linguistics: NAACL 2024},
  pages={3605--3627},
  year={2024}
}

@inproceedings{wang2025characterbox,
  title={Characterbox: Evaluating the role-playing capabilities of llms in text-based virtual worlds},
  author={Wang, Lei and Lian, Jianxun and Huang, Yi and Dai, Yanqi and Li, Haoxuan and Chen, Xu and Xie, Xing and Wen, Ji-Rong},
  booktitle={Proceedings of the 2025 Conference of the Nations of the Americas Chapter of the Association for Computational Linguistics: Human Language Technologies (Volume 1: Long Papers)},
  pages={6372--6391},
  year={2025}
}

@article{liu2024roleagent,
  title={Roleagent: Building, interacting, and benchmarking high-quality role-playing agents from scripts},
  author={Liu, Jiaheng and Ni, Zehao and Que, Haoran and Sun, Tao and Wang, Zekun and Yang, Jian and Wang, Jiakai and Guo, Hongcheng and Peng, Zhongyuan and Zhang, Ge and others},
  journal={Advances in Neural Information Processing Systems},
  volume={37},
  pages={49403--49428},
  year={2024}
}

@inproceedings{ran2025bookworld,
  title={BOOKWORLD: From Novels to Interactive Agent Societies for Story Creation},
  author={Ran, Yiting and Wang, Xintao and Qiu, Tian and Liang, Jiaqing and Xiao, Yanghua and Yang, Deqing},
  booktitle={Proceedings of the 63rd Annual Meeting of the Association for Computational Linguistics (Volume 1: Long Papers)},
  pages={15898--15912},
  year={2025}
}

@inproceedings{shao-etal-2023-character,
    title = "Character-{LLM}: A Trainable Agent for Role-Playing",
    author = "Shao, Yunfan  and
      Li, Linyang  and
      Dai, Junqi  and
      Qiu, Xipeng",
    editor = "Bouamor, Houda  and
      Pino, Juan  and
      Bali, Kalika",
    booktitle = "Proceedings of the 2023 Conference on Empirical Methods in Natural Language Processing",
    month = dec,
    year = "2023",
    address = "Singapore",
    publisher = "Association for Computational Linguistics",
    url = "https://aclanthology.org/2023.emnlp-main.814/",
    pages = "13153--13187"
}

@inproceedings{wang2024rolellm,
  title={Rolellm: Benchmarking, eliciting, and enhancing role-playing abilities of large language models},
  author={Wang, Noah and Peng, Zy and Que, Haoran and Liu, Jiaheng and Zhou, Wangchunshu and Wu, Yuhan and Guo, Hongcheng and Gan, Ruitong and Ni, Zehao and Yang, Jian and others},
  booktitle={Findings of the Association for Computational Linguistics: ACL 2024},
  pages={14743--14777},
  year={2024}
}

@inproceedings{chen2024agentverse,
  title={Agentverse: Facilitating multi-agent collaboration and exploring emergent behaviors},
  author={Chen, Weize and Su, Yusheng and Zuo, Jingwei and Yang, Cheng and Yuan, Chenfei and Chan, Chi-Min and Yu, Heyang and Lu, Yaxi and Hung, Yi-Hsin and Qian, Chen and others},
  booktitle={International Conference on Learning Representations},
  volume={2024},
  pages={20094--20136},
  year={2024}
}

@inproceedings{
qian2024merge,
title={''Merge Conflicts!''' Exploring the Impacts of External Knowledge Distractors to Parametric Knowledge Graphs},
author={Cheng Qian and Xinran Zhao and Tongshuang Wu},
booktitle={First Conference on Language Modeling},
year={2024},
url={https://openreview.net/forum?id=Pvn1dKreZW}
}

@inproceedings{xu-etal-2024-knowledge-conflicts,
    title = "Knowledge Conflicts for {LLM}s: A Survey",
    author = "Xu, Rongwu  and
      Qi, Zehan  and
      Guo, Zhijiang  and
      Wang, Cunxiang  and
      Wang, Hongru  and
      Zhang, Yue  and
      Xu, Wei",
    editor = "Al-Onaizan, Yaser  and
      Bansal, Mohit  and
      Chen, Yun-Nung",
    booktitle = "Proceedings of the 2024 Conference on Empirical Methods in Natural Language Processing",
    month = nov,
    year = "2024",
    address = "Miami, Florida, USA",
    publisher = "Association for Computational Linguistics",
    url = "https://aclanthology.org/2024.emnlp-main.486/",
    doi = "10.18653/v1/2024.emnlp-main.486",
    pages = "8541--8565"
}

@inproceedings{jain-wallace-2019-attention,
    title = "{A}ttention is not {E}xplanation",
    author = "Jain, Sarthak  and
      Wallace, Byron C.",
    editor = "Burstein, Jill  and
      Doran, Christy  and
      Solorio, Thamar",
    booktitle = "Proceedings of the 2019 Conference of the North {A}merican Chapter of the Association for Computational Linguistics: Human Language Technologies, Volume 1 (Long and Short Papers)",
    month = jun,
    year = "2019",
    address = "Minneapolis, Minnesota",
    publisher = "Association for Computational Linguistics",
    url = "https://aclanthology.org/N19-1357/",
    doi = "10.18653/v1/N19-1357",
    pages = "3543--3556"
}

@inproceedings{abnar-zuidema-2020-attention-flow,
    title = "Quantifying Attention Flow in Transformers",
    author = "Abnar, Samira  and
      Zuidema, Willem",
    editor = "Jurafsky, Dan  and
      Chai, Joyce  and
      Schluter, Natalie  and
      Tetreault, Joel",
    booktitle = "Proceedings of the 58th Annual Meeting of the Association for Computational Linguistics",
    month = jul,
    year = "2020",
    address = "Online",
    publisher = "Association for Computational Linguistics",
    url = "https://aclanthology.org/2020.acl-main.385/",
    doi = "10.18653/v1/2020.acl-main.385",
    pages = "4190--4197"
}

@inproceedings{tenney-etal-2019-bert-pipeline,
    title = "{BERT} Rediscovers the Classical {NLP} Pipeline",
    author = "Tenney, Ian  and
      Das, Dipanjan  and
      Pavlick, Ellie",
    editor = "Korhonen, Anna  and
      Traum, David  and
      M{\`a}rquez, Llu{\'i}s",
    booktitle = "Proceedings of the 57th Annual Meeting of the Association for Computational Linguistics",
    month = jul,
    year = "2019",
    address = "Florence, Italy",
    publisher = "Association for Computational Linguistics",
    url = "https://aclanthology.org/P19-1452/",
    doi = "10.18653/v1/P19-1452",
    pages = "4593--4601"
}

@inproceedings{chefer-etal-2021-beyond-attention,
  title     = {Transformer Interpretability Beyond Attention Visualization},
  author    = {Chefer, Hila and Gur, Shir and Wolf, Lior},
  booktitle = {Proceedings of the IEEE/CVF Conference on Computer Vision and Pattern Recognition (CVPR)},
  year      = {2021},
  pages     = {782--791},
  doi       = {10.1109/CVPR46437.2021.00084},
  url       = {https://openaccess.thecvf.com/content/CVPR2021/html/Chefer_Transformer_Interpretability_Beyond_Attention_Visualization_CVPR_2021_paper.html}
}

@inproceedings{yuan-etal-2024-evaluating,
    title = "Evaluating Character Understanding of Large Language Models via Character Profiling from Fictional Works",
    author = "Yuan, Xinfeng  and
      Yuan, Siyu  and
      Cui, Yuhan  and
      Lin, Tianhe  and
      Wang, Xintao  and
      Xu, Rui  and
      Chen, Jiangjie  and
      Yang, Deqing",
    editor = "Al-Onaizan, Yaser  and
      Bansal, Mohit  and
      Chen, Yun-Nung",
    booktitle = "Proceedings of the 2024 Conference on Empirical Methods in Natural Language Processing",
    month = nov,
    year = "2024",
    address = "Miami, Florida, USA",
    publisher = "Association for Computational Linguistics",
    url = "https://aclanthology.org/2024.emnlp-main.456/",
    doi = "10.18653/v1/2024.emnlp-main.456",
    pages = "8015--8036"
}

@inproceedings{NEURIPS2024_370df50c,
 author = {Penedo, Guilherme and Kydl\'{\i}\v{c}ek, Hynek and allal, Loubna Ben and Lozhkov, Anton and Mitchell, Margaret and Raffel, Colin and Von Werra, Leandro and Wolf, Thomas},
 booktitle = {Advances in Neural Information Processing Systems},
 doi = {10.52202/079017-0970},
 editor = {A. Globerson and L. Mackey and D. Belgrave and A. Fan and U. Paquet and J. Tomczak and C. Zhang},
 pages = {30811--30849},
 publisher = {Curran Associates, Inc.},
 title = {The FineWeb Datasets: Decanting the Web for the Finest Text Data at Scale},
 url = {https://proceedings.neurips.cc/paper_files/paper/2024/file/370df50ccfdf8bde18f8f9c2d9151bda-Paper-Datasets_and_Benchmarks_Track.pdf},
 volume = {37},
 year = {2024}
}

@inproceedings{wu-etal-2025-personas,
    title = "From Personas to Talks: Revisiting the Impact of Personas on {LLM}-Synthesized Emotional Support Conversations",
    author = "Wu, Shenghan  and
      Zhu, Yimo  and
      Hsu, Wynne  and
      Lee, Mong-Li  and
      Deng, Yang",
    editor = "Christodoulopoulos, Christos  and
      Chakraborty, Tanmoy  and
      Rose, Carolyn  and
      Peng, Violet",
    booktitle = "Proceedings of the 2025 Conference on Empirical Methods in Natural Language Processing",
    month = nov,
    year = "2025",
    address = "Suzhou, China",
    publisher = "Association for Computational Linguistics",
    url = "https://aclanthology.org/2025.emnlp-main.277/",
    doi = "10.18653/v1/2025.emnlp-main.277",
    pages = "5439--5453",
    ISBN = "979-8-89176-332-6"
}

@inproceedings{jun-lee-2025-exploring,
    title = "Exploring Persona Sentiment Sensitivity in Personalized Dialogue Generation",
    author = "Jun, Yonghyun  and
      Lee, Hwanhee",
    editor = "Che, Wanxiang  and
      Nabende, Joyce  and
      Shutova, Ekaterina  and
      Pilehvar, Mohammad Taher",
    booktitle = "Proceedings of the 63rd Annual Meeting of the Association for Computational Linguistics (Volume 1: Long Papers)",
    month = jul,
    year = "2025",
    address = "Vienna, Austria",
    publisher = "Association for Computational Linguistics",
    url = "https://aclanthology.org/2025.acl-long.900/",
    doi = "10.18653/v1/2025.acl-long.900",
    pages = "18384--18402",
    ISBN = "979-8-89176-251-0"
}

@article{yang2025qwen3,
  title={Qwen3 technical report},
  author={Yang, An and Li, Anfeng and Yang, Baosong and Zhang, Beichen and Hui, Binyuan and Zheng, Bo and Yu, Bowen and Gao, Chang and Huang, Chengen and Lv, Chenxu and others},
  journal={arXiv preprint arXiv:2505.09388},
  year={2025}
}

@article{agarwal2025gpt,
  title={gpt-oss-120b \& gpt-oss-20b model card},
  author={Agarwal, Sandhini and Ahmad, Lama and Ai, Jason and Altman, Sam and Applebaum, Andy and Arbus, Edwin and Arora, Rahul K and Bai, Yu and Baker, Bowen and Bao, Haiming and others},
  journal={arXiv preprint arXiv:2508.10925},
  year={2025}
}

@article{liu2025deepseek,
  title={Deepseek-v3. 2: Pushing the frontier of open large language models},
  author={Liu, Aixin and Mei, Aoxue and Lin, Bangcai and Xue, Bing and Wang, Bingxuan and Xu, Bingzheng and Wu, Bochao and Zhang, Bowei and Lin, Chaofan and Dong, Chen and others},
  journal={arXiv preprint arXiv:2512.02556},
  year={2025}
}

@article{comanici2025gemini,
  title={Gemini 2.5: Pushing the frontier with advanced reasoning, multimodality, long context, and next generation agentic capabilities},
  author={Comanici, Gheorghe and Bieber, Eric and Schaekermann, Mike and Pasupat, Ice and Sachdeva, Noveen and Dhillon, Inderjit and Blistein, Marcel and Ram, Ori and Zhang, Dan and Rosen, Evan and others},
  journal={arXiv preprint arXiv:2507.06261},
  year={2025}
}

@article{hurst2024gpt,
  title={Gpt-4o system card},
  author={Hurst, Aaron and Lerer, Adam and Goucher, Adam P and Perelman, Adam and Ramesh, Aditya and Clark, Aidan and Ostrow, AJ and Welihinda, Akila and Hayes, Alan and Radford, Alec and others},
  journal={arXiv preprint arXiv:2410.21276},
  year={2024}
}

@inproceedings{10.1145/3586183.3606763,
author = {Park, Joon Sung and O'Brien, Joseph and Cai, Carrie Jun and Morris, Meredith Ringel and Liang, Percy and Bernstein, Michael S.},
title = {Generative Agents: Interactive Simulacra of Human Behavior},
year = {2023},
isbn = {9798400701320},
publisher = {Association for Computing Machinery},
address = {New York, NY, USA},
url = {https://doi.org/10.1145/3586183.3606763},
doi = {10.1145/3586183.3606763},
booktitle = {Proceedings of the 36th Annual ACM Symposium on User Interface Software and Technology},
articleno = {2},
numpages = {22},
location = {San Francisco, CA, USA},
series = {UIST '23}
}

@inproceedings{chuang2024simulating,
  title={Simulating opinion dynamics with networks of llm-based agents},
  author={Chuang, Yun-Shiuan and Goyal, Agam and Harlalka, Nikunj and Suresh, Siddharth and Hawkins, Robert and Yang, Sijia and Shah, Dhavan and Hu, Junjie and Rogers, Timothy},
  booktitle={Findings of the association for computational linguistics: NAACL 2024},
  pages={3326--3346},
  year={2024}
}

@inproceedings{lu-etal-2024-large,
    title = "Large Language Models are Superpositions of All Characters: Attaining Arbitrary Role-play via Self-Alignment",
    author = "Lu, Keming  and
      Yu, Bowen  and
      Zhou, Chang  and
      Zhou, Jingren",
    editor = "Ku, Lun-Wei  and
      Martins, Andre  and
      Srikumar, Vivek",
    booktitle = "Proceedings of the 62nd Annual Meeting of the Association for Computational Linguistics (Volume 1: Long Papers)",
    month = aug,
    year = "2024",
    address = "Bangkok, Thailand",
    publisher = "Association for Computational Linguistics",
    url = "https://aclanthology.org/2024.acl-long.423/",
    doi = "10.18653/v1/2024.acl-long.423",
    pages = "7828--7840"
}

@misc{anthropic_claude_sonnet_45_2025,
  author       = {Anthropic},
  title        = {Introducing Claude Sonnet 4.5},
  year         = {2025},
  month        = sep,
  url          = {https://www.anthropic.com/news/claude-sonnet-4-5},
}

@article{samuel2024personagym,
  title={Personagym: Evaluating persona agents and llms},
  author={Samuel, Vinay and Zou, Henry Peng and Zhou, Yue and Chaudhari, Shreyas and Kalyan, Ashwin and Rajpurohit, Tanmay and Deshpande, Ameet and Narasimhan, Karthik and Murahari, Vishvak},
  journal={arXiv preprint arXiv:2407.18416},
  year={2024}
}

@inproceedings{
wang2025coser,
title={Co{SER}: Coordinating {LLM}-Based Persona Simulation of Established Roles},
author={Xintao Wang and Heng Wang and Yifei Zhang and Xinfeng Yuan and Rui Xu and Jen-tse Huang and Siyu Yuan and Haoran Guo and Jiangjie Chen and Shuchang Zhou and Wei Wang and Yanghua Xiao},
booktitle={Forty-second International Conference on Machine Learning},
year={2025},
url={https://openreview.net/forum?id=BOrR7YqKUt}
}

@inproceedings{wang-etal-2024-incharacter,
    title = "{I}n{C}haracter: Evaluating Personality Fidelity in Role-Playing Agents through Psychological Interviews",
    author = "Wang, Xintao  and
      Xiao, Yunze  and
      Huang, Jen-tse  and
      Yuan, Siyu  and
      Xu, Rui  and
      Guo, Haoran  and
      Tu, Quan  and
      Fei, Yaying  and
      Leng, Ziang  and
      Wang, Wei  and
      Chen, Jiangjie  and
      Li, Cheng  and
      Xiao, Yanghua",
    editor = "Ku, Lun-Wei  and
      Martins, Andre  and
      Srikumar, Vivek",
    booktitle = "Proceedings of the 62nd Annual Meeting of the Association for Computational Linguistics (Volume 1: Long Papers)",
    month = aug,
    year = "2024",
    address = "Bangkok, Thailand",
    publisher = "Association for Computational Linguistics",
    url = "https://aclanthology.org/2024.acl-long.102/",
    doi = "10.18653/v1/2024.acl-long.102",
    pages = "1840--1873"
}

@article{li2023chatharuhi,
  title={Chatharuhi: Reviving anime character in reality via large language model},
  author={Li, Cheng and Leng, Ziang and Yan, Chenxi and Shen, Junyi and Wang, Hao and Mi, Weishi and Fei, Yaying and Feng, Xiaoyang and Yan, Song and Wang, HaoSheng and others},
  journal={arXiv preprint arXiv:2308.09597},
  year={2023}
}

@inproceedings{huang2024emotional,
  title={Emotional RAG: Enhancing role-playing agents through emotional retrieval},
  author={Huang, Le and Lan, Hengzhi and Sun, Zijun and Shi, Chuan and Bai, Ting},
  booktitle={2024 IEEE International Conference on Knowledge Graph (ICKG)},
  pages={120--127},
  year={2024},
  organization={IEEE}
}

@inproceedings{sun2025identity,
  title={Identity-driven hierarchical role-playing agents},
  author={Sun, Libo and Wang, Siyuan and Wei, Zhongyu},
  booktitle={CCF International Conference on Natural Language Processing and Chinese Computing},
  pages={403--417},
  year={2025},
  organization={Springer}
}

@inproceedings{zhou2024characterglm,
  title={CharacterGLM: Customizing Social Characters with Large Language Models},
  author={Zhou, Jinfeng and Chen, Zhuang and Wan, Dazhen and Wen, Bosi and Song, Yi and Yu, Jifan and Huang, Yongkang and Ke, Pei and Bi, Guanqun and Peng, Libiao and others},
  booktitle={Proceedings of the 2024 Conference on Empirical Methods in Natural Language Processing: Industry Track},
  pages={1457--1476},
  year={2024}
}

@inproceedings{sadeq-etal-2024-mitigating,
    title = "Mitigating Hallucination in Fictional Character Role-Play",
    author = "Sadeq, Nafis  and
      Xie, Zhouhang  and
      Kang, Byungkyu  and
      Lamba, Prarit  and
      Gao, Xiang  and
      McAuley, Julian",
    editor = "Al-Onaizan, Yaser  and
      Bansal, Mohit  and
      Chen, Yun-Nung",
    booktitle = "Findings of the Association for Computational Linguistics: EMNLP 2024",
    month = nov,
    year = "2024",
    address = "Miami, Florida, USA",
    publisher = "Association for Computational Linguistics",
    url = "https://aclanthology.org/2024.findings-emnlp.846/",
    doi = "10.18653/v1/2024.findings-emnlp.846",
    pages = "14467--14479"
}

@inproceedings{zhang-etal-2025-revealing,
    title = "Revealing and Mitigating the Challenge of Detecting Character Knowledge Errors in {LLM} Role-Playing",
    author = "Zhang, Wenyuan  and
      Nie, Shuaiyi  and
      Sheng, Jiawei  and
      Zhang, Zefeng  and
      Zhang, Xinghua  and
      He, Yongquan  and
      Liu, Tingwen",
    editor = "Christodoulopoulos, Christos  and
      Chakraborty, Tanmoy  and
      Rose, Carolyn  and
      Peng, Violet",
    booktitle = "Proceedings of the 2025 Conference on Empirical Methods in Natural Language Processing",
    month = nov,
    year = "2025",
    address = "Suzhou, China",
    publisher = "Association for Computational Linguistics",
    url = "https://aclanthology.org/2025.emnlp-main.1689/",
    doi = "10.18653/v1/2025.emnlp-main.1689",
    pages = "33267--33290",
    ISBN = "979-8-89176-332-6"
}

@inproceedings{ahn-etal-2024-timechara,
    title = "{T}ime{C}hara: Evaluating Point-in-Time Character Hallucination of Role-Playing Large Language Models",
    author = "Ahn, Jaewoo  and
      Lee, Taehyun  and
      Lim, Junyoung  and
      Kim, Jin-Hwa  and
      Yun, Sangdoo  and
      Lee, Hwaran  and
      Kim, Gunhee",
    editor = "Ku, Lun-Wei  and
      Martins, Andre  and
      Srikumar, Vivek",
    booktitle = "Findings of the Association for Computational Linguistics: ACL 2024",
    month = aug,
    year = "2024",
    address = "Bangkok, Thailand",
    publisher = "Association for Computational Linguistics",
    url = "https://aclanthology.org/2024.findings-acl.197/",
    doi = "10.18653/v1/2024.findings-acl.197",
    pages = "3291--3325"
}

@inproceedings{yang-etal-2025-crafting,
    title = "Crafting Customisable Characters with {LLM}s: A Persona-Driven Role-Playing Agent Framework",
    author = "Yang, Bohao  and
      Liu, Dong  and
      Xiao, Chenghao  and
      Zhao, Kun  and
      Tang, Chen  and
      Li, Chao  and
      Yuan, Lin  and
      Guang, Yang  and
      Lin, Chenghua",
    editor = "Christodoulopoulos, Christos  and
      Chakraborty, Tanmoy  and
      Rose, Carolyn  and
      Peng, Violet",
    booktitle = "Findings of the Association for Computational Linguistics: EMNLP 2025",
    month = nov,
    year = "2025",
    address = "Suzhou, China",
    publisher = "Association for Computational Linguistics",
    url = "https://aclanthology.org/2025.findings-emnlp.1100/",
    doi = "10.18653/v1/2025.findings-emnlp.1100",
    pages = "20216--20240",
    ISBN = "979-8-89176-335-7"
}

@article{john1991big,
  title={Big five inventory},
  author={John, Oliver P and Donahue, Eileen M and Kentle, Robert L},
  journal={Journal of personality and social psychology},
  year={1991}
}

@article{
wang2024voyager,
title={Voyager: An Open-Ended Embodied Agent with Large Language Models},
author={Guanzhi Wang and Yuqi Xie and Yunfan Jiang and Ajay Mandlekar and Chaowei Xiao and Yuke Zhu and Linxi Fan and Anima Anandkumar},
journal={Transactions on Machine Learning Research},
issn={2835-8856},
year={2024},
url={https://openreview.net/forum?id=ehfRiF0R3a},
note={}
}

@article{mcadams2006new,
  title={A new Big Five: fundamental principles for an integrative science of personality.},
  author={McAdams, Dan P and Pals, Jennifer L},
  journal={American psychologist},
  volume={61},
  number={3},
  pages={204},
  year={2006},
  publisher={American Psychological Association}
}

@article{mischel1995cognitive,
  title={A cognitive-affective system theory of personality: reconceptualizing situations, dispositions, dynamics, and invariance in personality structure.},
  author={Mischel, Walter and Shoda, Yuichi},
  journal={Psychological review},
  volume={102},
  number={2},
  pages={246},
  year={1995},
  publisher={American Psychological Association}
}

@inproceedings{NEURIPS2022_b1efde53,
 author = {Ouyang, Long and Wu, Jeffrey and Jiang, Xu and Almeida, Diogo and Wainwright, Carroll and Mishkin, Pamela and Zhang, Chong and Agarwal, Sandhini and Slama, Katarina and Ray, Alex and Schulman, John and Hilton, Jacob and Kelton, Fraser and Miller, Luke and Simens, Maddie and Askell, Amanda and Welinder, Peter and Christiano, Paul F and Leike, Jan and Lowe, Ryan},
 booktitle = {Advances in Neural Information Processing Systems},
 editor = {S. Koyejo and S. Mohamed and A. Agarwal and D. Belgrave and K. Cho and A. Oh},
 pages = {27730--27744},
 publisher = {Curran Associates, Inc.},
 title = {Training language models to follow instructions with human feedback},
 url = {https://proceedings.neurips.cc/paper_files/paper/2022/file/b1efde53be364a73914f58805a001731-Paper-Conference.pdf},
 volume = {35},
 year = {2022}
}

@inproceedings{sheng2023flexgen,
  title={Flexgen: High-throughput generative inference of large language models with a single gpu},
  author={Sheng, Ying and Zheng, Lianmin and Yuan, Binhang and Li, Zhuohan and Ryabinin, Max and Chen, Beidi and Liang, Percy and R{\'e}, Christopher and Stoica, Ion and Zhang, Ce},
  booktitle={International Conference on Machine Learning},
  pages={31094--31116},
  year={2023},
  organization={PMLR}
}

@article{10.1145/3759441.3759444,
author = {Agrawal, Arney and Kedia, Nitin and Panwar, Ashish and Mohan, Jayashree and Kwatra, Nipun and Gulavani, Bhargav S. and Tumanov, Alexey and Ramjee, Ramachandran},
title = {Efficient LLM Inference via Chunked Prefills},
year = {2025},
issue_date = {July 2025},
publisher = {Association for Computing Machinery},
address = {New York, NY, USA},
volume = {59},
number = {1},
issn = {0163-5980},
url = {https://doi.org/10.1145/3759441.3759444},
doi = {10.1145/3759441.3759444},
journal = {SIGOPS Oper. Syst. Rev.},
month = aug,
pages = {9–16},
numpages = {8}
}

@article{liu2026ministral,
  title={Ministral 3},
  author={Liu, Alexander H and Khandelwal, Kartik and Subramanian, Sandeep and Jouault, Victor and Rastogi, Abhinav and Sad{\'e}, Adrien and Jeffares, Alan and Jiang, Albert and Cahill, Alexandre and Gavaudan, Alexandre and others},
  journal={arXiv preprint arXiv:2601.08584},
  year={2026}
}

@article{maaten2008visualizing,
  title={Visualizing data using t-SNE},
  author={Maaten, Laurens van der and Hinton, Geoffrey},
  journal={Journal of machine learning research},
  volume={9},
  number={Nov},
  pages={2579--2605},
  year={2008}
}

@article{lee2025gemini,
  title={Gemini embedding: Generalizable embeddings from gemini},
  author={Lee, Jinhyuk and Chen, Feiyang and Dua, Sahil and Cer, Daniel and Shanbhogue, Madhuri and Naim, Iftekhar and {\'A}brego, Gustavo Hern{\'a}ndez and Li, Zhe and Chen, Kaifeng and Vera, Henrique Schechter and others},
  journal={arXiv preprint arXiv:2503.07891},
  year={2025}
}

@inproceedings{li-etal-2023-contrastive,
    title = "Contrastive Decoding: Open-ended Text Generation as Optimization",
    author = "Li, Xiang Lisa  and
      Holtzman, Ari  and
      Fried, Daniel  and
      Liang, Percy  and
      Eisner, Jason  and
      Hashimoto, Tatsunori  and
      Zettlemoyer, Luke  and
      Lewis, Mike",
    editor = "Rogers, Anna  and
      Boyd-Graber, Jordan  and
      Okazaki, Naoaki",
    booktitle = "Proceedings of the 61st Annual Meeting of the Association for Computational Linguistics (Volume 1: Long Papers)",
    month = jul,
    year = "2023",
    address = "Toronto, Canada",
    publisher = "Association for Computational Linguistics",
    url = "https://aclanthology.org/2023.acl-long.687/",
    doi = "10.18653/v1/2023.acl-long.687",
    pages = "12286--12312"
}

@article{dong2026steer,
  title={Steer Model beyond Assistant: Controlling System Prompt Strength via Contrastive Decoding},
  author={Dong, Yijiang River and Hu, Tiancheng and Hui, Zheng and Collier, Nigel},
  journal={arXiv preprint arXiv:2601.06403},
  year={2026}
}

@inproceedings{preniqi2024moralbert,
  title={MoralBERT: a fine-tuned language model for capturing moral values in social discussions},
  author={Preniqi, Vjosa and Ghinassi, Iacopo and Ive, Julia and Saitis, Charalampos and Kalimeri, Kyriaki},
  booktitle={Proceedings of the 2024 International Conference on Information Technology for Social Good},
  pages={433--442},
  year={2024}
}

@article{zhao2023survey,
  title={A survey of large language models},
  author={Zhao, Wayne Xin and Zhou, Kun and Li, Junyi and Tang, Tianyi and Wang, Xiaolei and Hou, Yupeng and Min, Yingqian and Zhang, Beichen and Zhang, Junjie and Dong, Zican and others},
  journal={arXiv preprint arXiv:2303.18223},
  volume={1},
  number={2},
  pages={1--124},
  year={2023}
}

@article{huang2025safety,
  title={Safety tax: Safety alignment makes your large reasoning models less reasonable},
  author={Huang, Tiansheng and Hu, Sihao and Ilhan, Fatih and Tekin, Selim Furkan and Yahn, Zachary and Xu, Yichang and Liu, Ling},
  journal={arXiv preprint arXiv:2503.00555},
  year={2025}
}

@article{olmo2025olmo,
  title={Olmo 3},
  author={Olmo, Team and Ettinger, Allyson and Bertsch, Amanda and Kuehl, Bailey and Graham, David and Heineman, David and Groeneveld, Dirk and Brahman, Faeze and Timbers, Finbarr and Ivison, Hamish and others},
  journal={arXiv preprint arXiv:2512.13961},
  year={2025}
}

@article{rafailov2023direct,
  title={Direct preference optimization: Your language model is secretly a reward model},
  author={Rafailov, Rafael and Sharma, Archit and Mitchell, Eric and Manning, Christopher D and Ermon, Stefano and Finn, Chelsea},
  journal={Advances in neural information processing systems},
  volume={36},
  pages={53728--53741},
  year={2023}
}

@article{bai2022training,
  title={Training a helpful and harmless assistant with reinforcement learning from human feedback},
  author={Bai, Yuntao and Jones, Andy and Ndousse, Kamal and Askell, Amanda and Chen, Anna and DasSarma, Nova and Drain, Dawn and Fort, Stanislav and Ganguli, Deep and Henighan, Tom and others},
  journal={arXiv preprint arXiv:2204.05862},
  year={2022}
}

@inproceedings{
lambert2025tulu,
title={Tulu 3: Pushing Frontiers in Open Language Model Post-Training},
author={Nathan Lambert and Jacob Morrison and Valentina Pyatkin and Shengyi Huang and Hamish Ivison and Faeze Brahman and Lester James Validad Miranda and Alisa Liu and Nouha Dziri and Xinxi Lyu and Yuling Gu and Saumya Malik and Victoria Graf and Jena D. Hwang and Jiangjiang Yang and Ronan Le Bras and Oyvind Tafjord and Christopher Wilhelm and Luca Soldaini and Noah A. Smith and Yizhong Wang and Pradeep Dasigi and Hannaneh Hajishirzi},
booktitle={Second Conference on Language Modeling},
year={2025},
url={https://openreview.net/forum?id=i1uGbfHHpH}
}

@inproceedings{sclar2024quantifying,
  title={Quantifying Language Models' Sensitivity to Spurious Features in Prompt Design or: How I learned to start worrying about prompt formatting},
  author={Sclar, Melanie and Choi, Yejin and Tsvetkov, Yulia and Suhr, Alane},
  booktitle={International Conference on Learning Representations},
  volume={2024},
  pages={25055--25083},
  year={2024}
}

@article{he2024does,
  title={Does prompt formatting have any impact on llm performance?},
  author={He, Jia and Rungta, Mukund and Koleczek, David and Sekhon, Arshdeep and Wang, Franklin X and Hasan, Sadid},
  journal={arXiv preprint arXiv:2411.10541},
  year={2024}
}

@inproceedings{yi2026too,
  title={Too good to be bad: On the failure of llms to role-play villains},
  author={Yi, Zihao and Jiang, Qingxuan and Ma, Ruotian and Chen, Xingyu and Yang, Qu and Wang, Mengru and Ye, Fanghua and Shen, Ying and Tu, Zhaopeng and Li, Xiaolong and others},
  booktitle={Findings of the Association for Computational Linguistics: ACL 2026},
  pages={5724--5734},
  year={2026}
}

@inproceedings{sharma2024towards,
  title={Towards understanding sycophancy in language models},
  author={Sharma, Mrinank and Tong, Meg and Korbak, Tomek and Duvenaud, David and Askell, Amanda and Bowman, Sam and Durmus, Esin and Hatfield-Dodds, Zac and Johnston, Scott and Kravec, Shauna and others},
  booktitle={International Conference on Learning Representations},
  volume={2024},
  pages={110--144},
  year={2024}
}

\clearpage
\appendix

\section{Appendix Table of Contents}
\label{app:toc}
We organize the Appendix as follows:

\begin{enumerate}[leftmargin=*, itemsep=2pt, topsep=2pt]
    \item Section~\ref{app:LLMs} provides the overall usage of Large Language Models (LLMs).
    
    \item Section~\ref{app:Dataset} provides dataset details, including the full character-profile schema, known and unknown character construction pipelines, moral--immoral disposition validation, unstructured-profile generation, and dataset validation.

    \item Section~\ref{app:RQ1_ExpSetup} describes the experimental setup, including model and API configurations, benchmark protocols for \textsc{PersonaGym} and \textsc{CoSER}, scenario generation, and implementation details for the evaluation pipeline.

    \item Section~\ref{app:human_eval} reports human evaluation results for both RQ1 and RQ2, validating the automatic evaluation trends for the main axis analysis and the FACD mitigation results.

    \item Section~\ref{app:RQ1_Detailed} provides detailed RQ1 analyses, including per-metric results for the Structure, Familiarity, and Disposition axes, the counterfactual rewrite prompt, and the full field-level localization results across models.

    \item Section~\ref{app:known_unknown_turn} further analyzes the Familiarity axis through turn-length ablations and attention/saturation diagnostics, showing how multi-turn interaction changes reliance on profile, history, and self-generated context.

    \item Section~\ref{app:facd_details} gives additional details on Field-Aware Contrastive Decoding, including field-level immoral filtering, a concrete Moriarty example, and the disposition-robust fallback fields used to construct the sanitized negative prompt.
\end{enumerate}

\section{Use of Large Language Models}
\label{app:LLMs}
We used large language models as assistance tools in several parts of the research process.
First, we used ChatGPT for improving the clarity of the manuscript; all substantive claims, analyses, and final wording were reviewed and edited by the authors.
Second, we used LLMs (Claude-sonnet-4.5, Claude-haiku-4.5, Gemini-2.5-pro) during the construction and processing of character profiles, including [profile generation / summarization / rewriting / filtering].
Third, we used GPT-4o as LLM-as-a-Judge for evaluating role-playing performance. We use same prompt templates described in prior works: PersonaGym~\cite{samuel2024personagym} and CoSER~\cite{wang2025coser}. We also manually inspected the outputs.
Finally, we used Claude Code as a coding assistant for implementation support, including code drafting, debugging, and refactoring. All generated or modified code was reviewed, tested, and verified by the authors.

\section{Dataset Details}
\label{app:Dataset}

\subsection{Dataset Schema}
\label{app:Dataset_Schema}
The dataset is structured using a hierarchical schema that represents character identity beyond personality traits alone, decomposing it into multiple high-level dimensions such as personal attributes, personality traits, relationships, motivations, and abilities.
Detailed field definitions and example schema templates are provided in Table~\ref{prompt:template1} and Table~\ref{prompt:template2}. For profile generation prompt, see Table~\ref{prompt:summarization}.

\subsection{Dataset Construction Details}
\subsubsection{Known Character Profiles}
\label{app:known_character}
We constructed character profiles for known characters using character metadata from Fandom. For each work, we extracted 3–5 characters, restricting our selection to widely recognized franchises and prioritizing high-salience characters with substantial narrative prominence. The resulting pool of selected works and the number of characters chosen per work are summarized in Table~\ref{prompt:works}.


\subsubsection{Unknown Character Profiles}
\label{app:unknown_character}
Unknown characters are constructed through a structured prompt-based pipeline that mirrors the schema and construction procedure used for known characters.
The pipeline consists of three components.

First, we define a profile skeleton that specifies the demographic attributes, personality trait seeds, and scenario genres from which unknown characters are instantiated.
This skeleton provides only structural constraints and candidate value pools, without enforcing narrative coherence.
The full schema and organization are shown in Table~\ref{prompt:synthetic sample}.

Second, to ensure internal logical consistency among sampled attributes, we employ an LLM-based profile validation prompt.
This judge prompt evaluates whether a sampled profile contains intrinsic conflicts (e.g., age--education--occupation or family status--age mismatches) and assigns a coherence score on a 1--10 scale.
Only profiles that satisfy basic logical coherence are retained.
The validation prompt is detailed in Table~\ref{prompt:character Validation}.

Finally, for downstream experiments requiring narrative context, we use a separate story generation prompt that conditions on the validated profile to produce short descriptive scenes.
This prompt is used solely to generate auxiliary narrative material and does not alter the underlying character attributes.
The corresponding prompt template is provided in Table~\ref{prompt:storyPrompt}.

\subsubsection{Verifying Disposition Classification with Human Validation}
\label{app:moral_immoral_details}
\begin{table}[h]
\centering
\resizebox{0.9\linewidth}{!}{%
\begin{tabular}{l|cc}
\toprule
\textbf{Fields} & \textbf{Agreement} & \textbf{$\kappa$} \\
\midrule
\texttt{Personality Traits} & 0.875 & 0.744 \\
\texttt{Interpersonal Relationships} & 0.9 & 0.8 \\
\texttt{Abilities} & 0.85 & 0.70 \\
\texttt{Motivations} & 0.95 & 0.89 \\
\midrule
\textit{Overall Profile} & \textbf{0.90} & \textbf{0.80} \\
\bottomrule
\end{tabular}%
}
\caption{LLM–human agreement and Cohen's $\kappa$ scores for Moral–Immoral disposition classification.}
\label{tab:profile_human}
\vspace{-2mm}
\end{table}
To validate the \textit{Moral}-\textit{Immoral} classification, we sample 20 \textit{Moral} and 20 \textit{Immoral} profiles and have three graduate students in an AI research lab with documented English proficiency annotators independently label them, determining the final label through majority voting. As shown in Table~\ref{tab:profile_human}, the agreement between the LLM and human classifications for the overall profiles reaches 0.90 with a Cohen's $\kappa$ of 0.80, confirming reliable disposition labeling across all profile dimensions. Furthermore, the agreement scores for the individual detailed fields consistently exceed 0.85, with corresponding $\kappa$ values all above 0.70, demonstrating substantial inter-annotator consistency. Also, we provide full classification prompt in Table~\ref{prompt:attributive_judge}.

\subsubsection{Unstructured Profiles}
\label{app:unstructured_details}
Based on the structured profiles we developed, we generated a parallel set of unstructured profiles. The detailed prompt used for this process is provided in Table~\ref{prompt:unstructure}.

\subsection{Dataset Validation}
\label{app:data_validation}

\begin{figure}[h]
    \centering
    \includegraphics[width=0.90\linewidth]{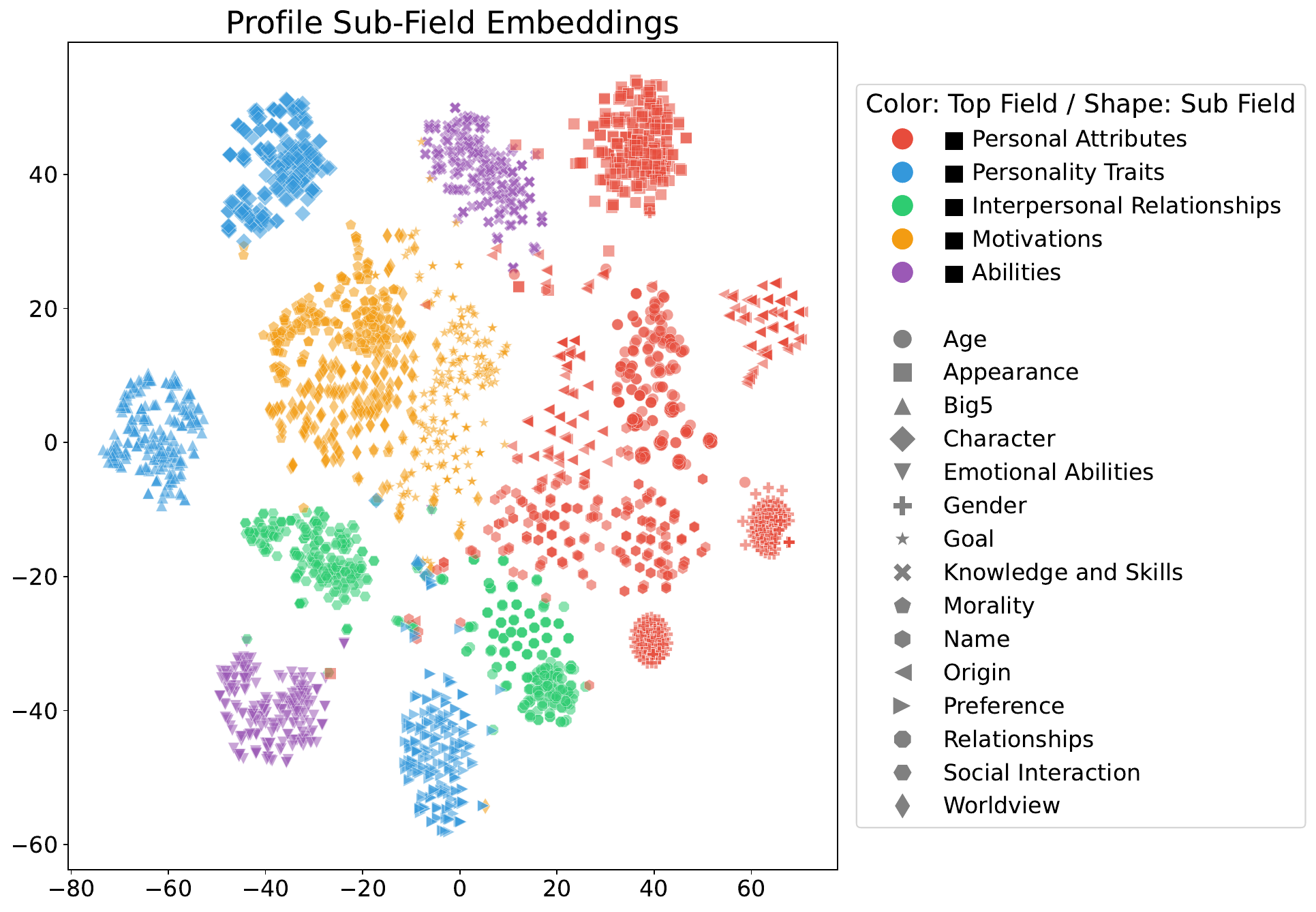}
    \vspace{-2mm}
    \caption{t-SNE visualization of profile sub-field embeddings, colored by dimension and shaped by leaf field.}
    \label{fig:embedding_cluster}
    \vspace{-3mm}
\end{figure}

\paragraph{Field Independence}
To verify each profile field captures distinct information, we compute Gemini-001 embeddings~\cite{lee2025gemini} for all leaf fields and visualize them via t-SNE~\cite{maaten2008visualizing}. As shown in Figure~\ref{fig:embedding_cluster}, fields cluster separately by dimension, confirming that the schema produces non-overlapping content across fields.

\paragraph{Profile Reliability}
To ensure the profiles accurately reflect the metadata without hallucination, we use GPT-4o to decompose each profile into atomic facts~\cite{sadeq-etal-2024-mitigating}. We then segment the metadata into chunks of up to 1,200 words. For each atomic fact, we retrieve the top-3 most semantically similar chunks and prompt Gemini-3.1-Pro to verify whether these evidence chunks support the given fact. Our analysis reveals that the profiles are decomposed into an average of 153.3 atomic facts, and 141.04 of these are explicitly supported by the metadata chunks, achieving a precision score of 0.92. This confirms that our profiles are highly faithful to the source material and possess robust reliability.

\section{Experimental Details}
\label{app:RQ1_ExpSetup}
Most models were accessed via the OpenRouter API with temperature set to 0 (greedy decoding) for role-playing agents and judge models. 
We used the following providers: TogetherAI for Qwen3 and GPT-oss families, and AtlasCloud for the Deepseek-v3.2 model. For mechanical analysis, we used 2 NVIDIA A6000 GPUs.

\subsection{Benchmark Details}
\paragraph{PersonaGym} \textsc{PersonaGym} evaluates persona agents using a deliberately \textit{single-character}, \textit{single-turn protocol}: each evaluation instance assigns exactly one persona (typically via a persona-conditioning system instruction) and asks one task question, and the agent produces one response; instances are scored independently, so the benchmark does not test multi-turn state tracking. In the full framework, an LLM “reasoner” first selects persona-relevant contexts from a pool of 150 environments, then generates task-specific questions for the chosen settings; the released static benchmark fixes this process into 200 personas and 10,000 questions for standardized comparison across models. 
\textsc{PersonaGym} covers five task dimensions—Action Justification, Expected Action, Linguistic Habits, Persona Consistency, and Toxicity Control—and scores each single-turn response with detailed, task-specific rubrics on a 1–5 ordinal scale, augmented with persona-and-question–conditioned exemplar responses to calibrate judging. Each response is graded by an ensemble of two evaluator LLMs, and the final score is computed by averaging the judges’ outputs; results are then aggregated across tasks into an overall \textit{PersonaScore}.

\paragraph{CoSER} \textsc{CoSER} evaluates role-playing language agents in multi-turn, multi-character literary scenes via Given-Circumstance Acting (GCA), where an “actor” LLM is required to sequentially portray multiple characters to reconstruct authentic scenarios extracted from well-known novels. 
Concretely, \textsc{CoSER} provides multi-party dialogues and rich contextual conditioning signals (e.g., scenario descriptions, character profiles, motivations, and—in the dataset representation—optional speech/action/thought annotations), and explicit support for >2-character interactions. 
At evaluation time, GCA first runs a multi-agent simulation: one RPA instance is created per character using the same actor LLM, each conditioned on the shared scenario plus its own persona materials, while a next-speaker-prediction (NSP) component selects who speaks next and a separate “environment” model emits environmental feedback; the rollout terminates either when NSP outputs an END signal or when a maximum turn budget (e.g., 18 turns) is reached. To increase controllability, GCA additionally applies penalty-based LLM judging: instead of asking a judge to output a single holistic score, a critic LLM identifies rubric-defined flaws (e.g., deviations from the reference dialogue or lack of initiative), assigns each flaw a severity level (1–5), aggregates penalties into per-dimension scores, and applies a length-correction procedure to mitigate bias toward conversation length. 
The rubric-driven scores are reported over three core dimensions: Anthropomorphism (the extent to which RPAs speak and act in a human-like manner), Character Fidelity (the degree to which the character profile is faithfully reflected in the generated narrative), and Storyline Quality (how natural and engaging the narrative flow is). In our setting, because our dataset instances do not provide explicit scenario descriptions, we generate synthetic scenarios using Gemini-2.5-Pro (temperature = 0.7), conditioning on the set of characters participating in each conversation. To mitigate potential bias induced by any single scenario formulation, we construct three clearly distinct seed scenarios per work and use these as the basis for interaction. For scenario generation prompt, see Table~\ref{prompt:scenario_generation}.

Notably, both benchmarks have demonstrated strong correlation with human judgment in their original studies.

\section{Human Evaluation}
\label{app:human_eval}
We conduct human evaluation on both PersonaGym and CoSER using two backbone models (Qwen3-235B and Mistral-Small). For each benchmark, we randomly sample 20\% of characters from each condition and partition them into three non-overlapping groups. Each of three annotators, all bachelor's degree holders with English proficiency, is assigned one group and evaluates the corresponding outputs from both models; thus, each sampled character--model pair is rated by one annotator. Annotators assign a single score on a five-point Likert scale to each sample using a unified rubric that considers both character fidelity and dialogue quality. Table~\ref{prompt:human_eval} provides the full evaluation instructions and score definitions. Results are reported in Table~\ref{tab:rq1human}. We reuse the same sampled character--model pairs for the FACD mitigation evaluation in Table~\ref{tab:rq2human}.

\subsection{RQ1 Human Validation Results}
\label{app:rq1_human}
\begin{table}[h]
    \centering
    \small
    \resizebox{0.9\columnwidth}{!}{%
    \begin{tabular}{@{}l|cc|cc}
    \toprule
        & \multicolumn{2}{c|}{\textbf{Familiarity}}& \multicolumn{2}{c}{\textbf{Disposition}} \\
         &  \textbf{Known} & \textbf{Unknown}  & \textbf{Moral} & \textbf{Immoral} \\
         \midrule
         \rowcolor{gray!30}
         \multicolumn{5}{c}{\textbf{PersonaGym (Single-turn interview)}} \\
    \midrule
         \textbf{Qwen3-8B} & 4.29 & 4.25 & 4.32 & 4.11\\
         
         \textbf{Mistral-Small} & 4.20 & 4.16 & 4.34 & 4.21  \\ 
         
         \midrule
    \rowcolor{gray!30}
    \multicolumn{5}{c}{\textbf{CoSER (Multi-turn interaction)}} \\
    \midrule
         \textbf{Qwen3-8B} & 3.00 & 2.89 & 3.19 & 2.69 \\
         
         \textbf{Mistral-Small} & 3.47 & 3.53 & 3.63 & 3.39 \\

         \bottomrule
    \end{tabular}%
    }
    \caption{Human evaluation scores for RQ1 across Familiarity and Disposition axes on both benchmarks.}
    \label{tab:rq1human}
\end{table}

As shown in Table~\ref{tab:rq1human}, human judgments are consistent with the LLM-based evaluation: the Disposition axis produces the largest performance gap, while Familiarity differences remain marginal.

\subsection{RQ2 Human Validation Results}
\label{app:rq2_human}
\begin{table}[h]
    \centering
    \small
    \begin{tabular}{@{}l|cc|c}
    \toprule
         & \textbf{Moral} & \textbf{Immoral} & \textit{Avg.} \\
    \midrule
    \rowcolor{gray!30}
    \multicolumn{4}{c}{\textbf{Default}} \\
    \midrule
         \textbf{Qwen3-8B} & 3.19 & 2.69  & 2.94 \\    
         \textbf{Mistral-Small} & 3.63 & 3.39 & 3.51 \\
    \midrule
    \rowcolor{pink!30}
    \multicolumn{4}{c}{\textbf{Field-Aware Contrastive Decoding}} \\
    \midrule
        \textbf{Qwen3-8B} & 3.56 & 3.25  & 3.40 \\    
         \textbf{Mistral-Small} & 3.68 & 3.47 & 3.58 \\
        
    \bottomrule
    \end{tabular}
    \caption{Human evaluation scores for FACD mitigation on CoSER.}
    \label{tab:rq2human}
\end{table}

Table~\ref{tab:rq2human} confirms that FACD improves Immoral character scores under human evaluation as well, aligning with the automatic evaluation trends reported in Section~\ref{sec:rq2_results}.

\section{RQ1 Detailed Analysis}
\label{app:RQ1_Detailed}
This section provides detailed per-metric results across all five models, further analyses, and used prompts examined in Section~\ref{sec:rq1}.

\subsection{Detailed Results per Axis}
\label{app:per_axis}

\paragraph{Structure Axis}
\label{app:str_unstr}
Per-metric breakdowns for the \textit{Structure} axis comparison are provided in Table~\ref{tab:struct_vs_unstruct}.

\paragraph{Familiarity Axis}
\label{app:known_unknown}
Per-metric breakdowns for the \textit{Familiarity} axis comparison are provided in Table~\ref{tab:known_vs_unknown}.

\paragraph{Disposition Axis}
\label{app:moral_immoral}
Per-metric breakdowns for the \textit{Disposition} axis comparison are provided in Table~\ref{tab:moral_vs_immoral}.

\subsection{Details for Counterfactual Rewriting}
\label{app:rewrite_prompt}

\begin{table}[t]
\centering
\small
\resizebox{\columnwidth}{!}{
\begin{tabular}{l|l|cc||cc}
\toprule
\multirow{2}{*}{\textbf{Model}} 
& \multirow{2}{*}{\textbf{Type}} 
& \multicolumn{2}{c||}{\textbf{Familiarity}} 
& \multicolumn{2}{c}{\textbf{Disposition}} \\
& & \textbf{\textit{K}} & \textbf{\textit{UK}} & \textbf{\textit{M}} & \textbf{\textit{IM}} \\
\midrule

\rowcolor{gray!30}
\multicolumn{6}{c}{\textbf{PersonaGym}} \\
\midrule

\multirow{3}{*}{Qwen3-235B} 
& \textit{Original} 
& 4.66 & 4.62 & 4.80 & 4.47 \\
& \textit{Counter}  
& 4.49 & 4.44 & 3.98 & 4.73 \\
\cmidrule(lr){2-6}
& $\Delta$ 
& -0.17
& -0.18 
& \textbf{-0.82} 
& \textbf{+0.26} \\
\midrule

\multirow{3}{*}{DeepSeek-v3.2} 
& \textit{Original}
& 4.53 & 4.54 & 4.68 & 4.39 \\
& \textit{Counter} 
& 4.37 & 4.43 & 4.06 & 4.73 \\
\cmidrule(lr){2-6}
& $\Delta$ 
& -0.16 
& -0.11 
& \textbf{-0.62} 
& \textbf{+0.34} \\

\midrule
\rowcolor{gray!30}
\multicolumn{6}{c}{\textbf{CoSER}} \\
\midrule

\multirow{3}{*}{Qwen3-235B} 
& \textit{Original}
& 33.30 & 35.00 & 38.06 & 29.66 \\
& \textit{Counter} 
& 32.28 & 35.81 & 15.89 & 33.13 \\
\cmidrule(lr){2-6}
& $\Delta$ 
& -1.02 
& +0.81 
& \textbf{-22.17} 
& \textbf{+3.47} \\
\midrule

\multirow{3}{*}{DeepSeek-v3.2} 
& \textit{Original} 
& 38.10 & 42.63 & 44.82 & 35.60 \\
& \textit{Counter}  
& 37.55 & 42.42 & 24.93 & 37.39 \\
\cmidrule(lr){2-6}
& $\Delta$ 
& -0.55 
& -0.21 
& \textbf{-19.89} 
& \textbf{+1.79} \\
\bottomrule
\end{tabular}
}
\vspace{-3mm}
\caption{Counterfactual rewrite results. K and UK denote \textit{Known} and \textit{Unknown}, respectively, while M and IM denote \textit{Moral} and \textit{Immoral}.}
\label{tab:counterfactual_rewrite}
\vspace{-5mm}
\end{table}

\newcommand{\increase}[1]{%
    \textcolor{red}{\scriptsize\,(+#1)}%
}
\newcommand{\decrease}[1]{%
    \textcolor{blue}{\scriptsize\,(-#1)}%
}

\begin{table}[t]
    \centering
    \small
    \setlength{\tabcolsep}{7pt}
    \renewcommand{\arraystretch}{1.12}

    \resizebox{\linewidth}{!}{%
        \begin{tabular}{l|ll|cc}
            \toprule
            \textbf{Model}
            & \textbf{Type}
            & \textbf{Field}
            & \textbf{\textit{M}}
            & \textbf{\textit{IM}} \\
            \midrule

            \rowcolor{gray!20}
            \multicolumn{5}{c}{\textbf{PersonaGym}} \\
            \midrule

            \multirow{4}{*}{Qwen3-235B}
            & \textit{Original}
            & Full
            & 4.80
            & 4.47 \\
            \cmidrule(lr){2-5}

            & \multirow{3}{*}{\textit{Counter}}
            & \textit{Motivations}
            & 4.36\decrease{0.44}
            & 4.60\increase{0.13} \\

            &
            & \textit{Morality}
            & 4.37\decrease{0.43}
            & 4.59\increase{0.12} \\

            &
            & \textit{Goal}
            & 4.18\decrease{0.62}
            & 4.57\increase{0.10} \\

            \midrule

            \multirow{4}{*}{DeepSeek-v3.2}
            & \textit{Original}
            & Full
            & 4.68
            & 4.39 \\
            \cmidrule(lr){2-5}

            & \multirow{3}{*}{\textit{Counter}}
            & \textit{Motivations}
            & 4.54\decrease{0.14}
            & 4.72\increase{0.33} \\

            &
            & \textit{Morality}
            & 4.55\decrease{0.13}
            & 4.71\increase{0.32} \\

            &
            & \textit{Goal}
            & 4.54\decrease{0.14}
            & 4.72\increase{0.33} \\

            \midrule

            \rowcolor{gray!20}
            \multicolumn{5}{c}{\textbf{CoSER}} \\
            \midrule

            \multirow{4}{*}{Qwen3-235B}
            & \textit{Original}
            & Full
            & 38.06
            & 29.66 \\
            \cmidrule(lr){2-5}

            & \multirow{3}{*}{\textit{Counter}}
            & \textit{Motivations}
            & 25.89\decrease{12.17}
            & 32.03\increase{2.37} \\

            &
            & \textit{Morality}
            & 26.51\decrease{11.55}
            & 31.89\increase{2.23} \\

            &
            & \textit{Goal}
            & 25.95\decrease{12.11}
            & 29.99\increase{0.33} \\

            \midrule

            \multirow{4}{*}{DeepSeek-v3.2}
            & \textit{Original}
            & Full
            & 44.82
            & 35.60 \\
            \cmidrule(lr){2-5}

            & \multirow{3}{*}{\textit{Counter}}
            & \textit{Motivations}
            & 29.29\decrease{15.53}
            & 37.39\increase{1.79} \\

            &
            & \textit{Morality}
            & 32.05\decrease{12.77}
            & 36.19\increase{0.59} \\

            &
            & \textit{Goal}
            & 30.12\decrease{14.70}
            & 36.75\increase{1.15} \\

            \bottomrule
        \end{tabular}%
    }

    \caption{
        Model performance on PersonaGym and CoSER under the original
        and counterfactual persona settings.
        Values in parentheses indicate changes relative to the corresponding
        \textit{Original} setting.
    }
    \label{tab:rewrite_disposition}
\end{table}

To examine whether the observed gaps follow the intended profile axes rather than fixed character identities, we conduct counterfactual rewrite experiments on shared base profiles. For the \textbf{Familiarity} axis, we pair each \textit{Known} profile with its embedding-nearest \textit{Unknown} profile and swap only the \texttt{Personal Attributes} field. The resulting counterfactual profile shares the same \texttt{Personal Attributes} while all remaining fields come from the paired profile. As shown in Table~\ref{tab:counterfactual_rewrite}, performance remains broadly stable after this identity swap. Although counterfactual profiles show a slight overall decline, likely due to conflicts between the swapped identity attributes and the remaining profile content, the magnitude of this decline remains limited. This further supports our finding that \textbf{Familiarity} is not a major bottleneck in role-playing performance.

For the \textbf{Disposition} axis, we first rewrite all non-\texttt{Personal Attributes} fields of each \textit{Moral} profile toward immoral content, and vice versa, using Claude-4.5-Haiku while preserving profile length, specificity, and writing style (full prompt is provided in Table~\ref{prompt:rewrite}). Unlike Familiarity, this rewrite produces clear directional shifts: \textit{Moral}$\rightarrow$\textit{Immoral} substantially lowers performance on both PersonaGym ($-0.82/-0.62$) and CoSER ($-22.17/-19.89$), whereas \textit{Immoral}$\rightarrow$\textit{Moral} improves performance on PersonaGym ($+0.26/+0.34$) and CoSER ($+3.47/+1.79$). These results provide additional evidence that the Disposition gap follows the rewritten dispositional content rather than the original character identity.

To more directly isolate the source of this effect, we further restrict the intervention to a single value-laden field while holding all other profile content fixed. As shown in Table~\ref{tab:rewrite_disposition}, rewriting only the \texttt{Motivations} field—or, more narrowly, its \texttt{Morality} or \texttt{Goal} subfields—reproduces the same directional pattern on both benchmarks: Moral-to-Immoral edits consistently degrade performance, whereas Immoral-to-Moral edits improve it. Because all other co-occurring profile content remains unchanged, these localized interventions more cleanly isolate the contribution of dispositional polarity itself.

\subsection{Additional RLVR-aligned checkpoint.}
\label{app:rlvr_ablation}
\begin{table}[t]
\centering
\small
\resizebox{0.8\columnwidth}{!}{
\begin{tabular}{l|cccc}
\toprule
\textbf{Rubrics}
& \textbf{\textit{M}} 
& \textbf{\textit{IM}}
& \textbf{$\Delta$}
& \textbf{\textit{p}} \\
\midrule

\rowcolor{gray!20}
\multicolumn{5}{c}{\textbf{Instruct-SFT}} \\
\midrule

\textit{Anthro.} 
& 10.78 & 6.02 & -4.76 & * \\
\textit{Char. Fid.} 
& 15.88 & 13.44 & -2.44 & - \\
\textit{Story Qual.} 
& 32.45 & 26.66 & -5.79 & * \\
\midrule
\textit{Avg.}
& 19.70 & 15.37 & -4.53 & * \\
\midrule

\rowcolor{red!15}
\multicolumn{5}{c}{\textbf{$+$ DPO}} \\
\midrule

\textit{Anthro.} 
& 13.53 & 6.32 & \textbf{-7.21} & *** \\
\textit{Char. Fid.} 
& 16.67 & 7.01 & \textbf{-9.66} & *** \\
\textit{Story Qual.} 
& 46.08 & 39.95 & -6.13 & ** \\
\midrule
\textit{Avg.}
& 25.42 & 17.79 & \textbf{-7.63} & *** \\

\midrule
\rowcolor{blue!15}
\multicolumn{5}{c}{\textbf{$+$ RLVR}} \\
\midrule

\textit{Anthro.} 
& 15.47 & 10.56 & -4.91 & * \\
\textit{Char. Fid.} 
& 15.24 & 7.58 & -7.66 & *** \\
\textit{Story Qual.} 
& 49.71 & 42.72 & \textbf{-6.99} & ** \\
\midrule
\textit{Avg.}
& 26.81 & 20.28 & -6.53 & *** \\

\bottomrule
\end{tabular}
}
\caption{Alignment-stage ablation on CoSER. \textit{M}, \textit{IM} denote \textit{Moral}, \textit{Immoral}; \textit{Anthro.}, \textit{Char. Fid.}, and \textit{Story Qual.} denote \textit{Anthropomorphism}, \textit{Character Fidelity}, and \textit{Storyline Quality}. Significance levels: * $p<0.05$, ** $p<0.01$, *** $p<0.001$.}
\label{tab:rlvr_ablation}
\vspace{-2mm}
\end{table}

As discussed in Section~\ref{sec:rq1_discussion}, DPO alignment widens the Moral--Immoral gap from $-4.53$ to $-7.63$, largely by degrading \textit{Immoral} \textit{Character Fidelity}. We further evaluate OLMo-3.1-32B-Instruct after RLVR alignment, which optimizes verifiable objectives such as math and coding~\cite{lambert2025tulu}. As shown in Table~\ref{tab:rlvr_ablation}, RLVR improves the average scores for both \textit{Moral} and \textit{Immoral} characters, suggesting incidental transfer of general reasoning or instruction-following ability to role-playing. However, it does not eliminate the DPO-amplified asymmetry: the average gap remains large and significant ($\Delta=-6.53$), and the \textit{Character Fidelity} gap remains much larger than under Instruct-SFT ($-7.66$ vs. $-2.44$).

\subsection{Detailed Localization Results}
\label{app:localization_full}
\begin{figure*}[!htbp]
    \centering
    \includegraphics[width=1\linewidth]{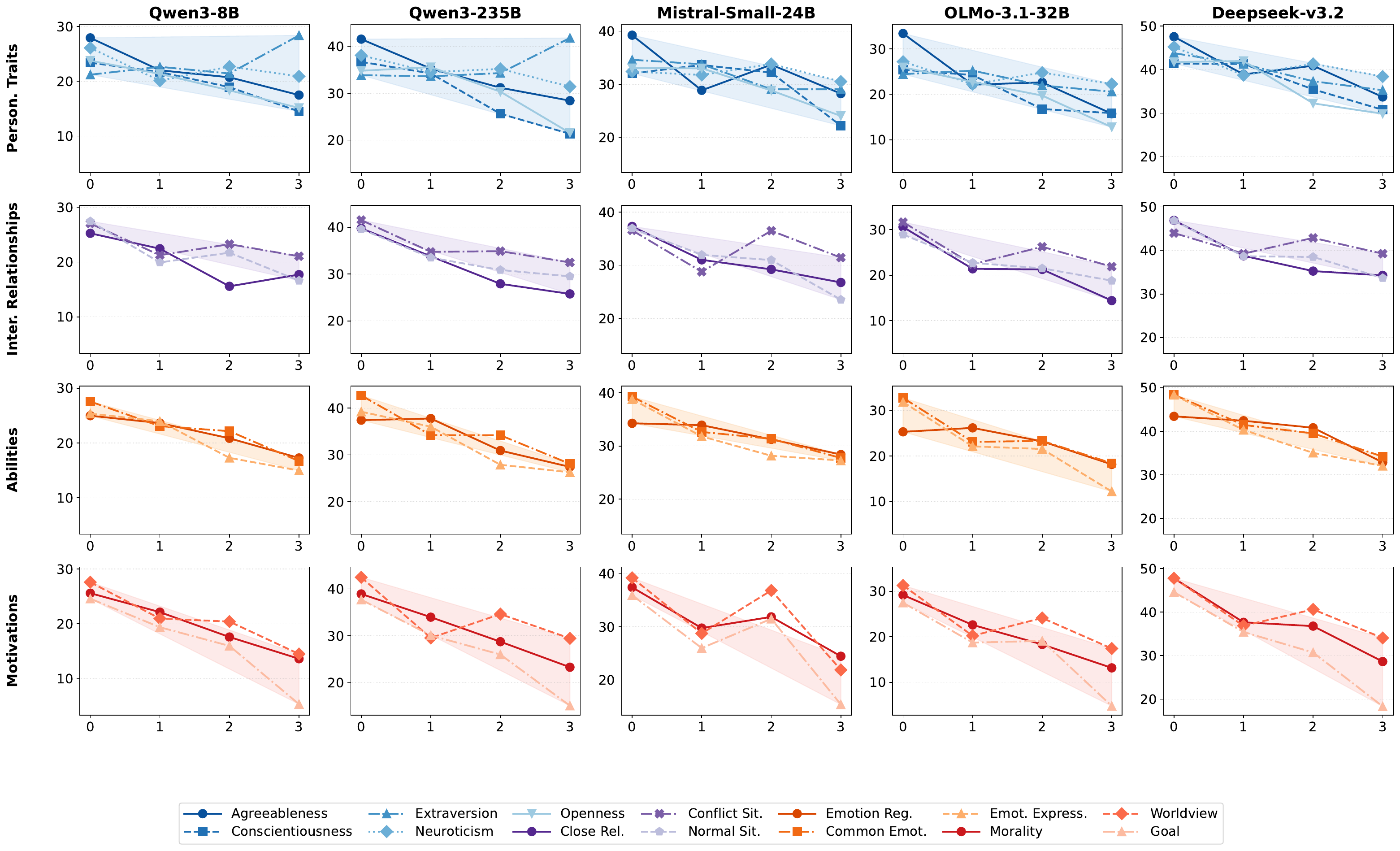}
    \vspace{-5mm}
    \caption{Impact of immoral character on CoSER scores. \textbf{(a) Personality Traits} exhibit relatively stable performance, while \textbf{(b) Motivations} show consistent decreasing trends as more immoral characters.}
    \label{fig:localization_full}
    \vspace{-5mm}
\end{figure*}

Figure~\ref{fig:localization_full} provides the full five-model version of the localization analysis shown in Figure~\ref{fig:rq2_Coser}.

\section{Detailed Analysis for Familiarity Axis}
\label{app:known_unknown_turn}

\subsection{Turn Ablation}
\subsubsection{Motivation and Setting}
Interacting environments differ from interviewing settings in agent count (single vs. multi) and duration (single vs. multi-turn). Inspired by findings that parametric knowledge can be detrimental in long-term contexts~\cite{qian2024merge, xu-etal-2024-knowledge-conflicts}, we hypothesize that the multi-turn factor suppresses the benefits of parametric identities, thereby bridging the gap between known and unknown characters. To test this, we track performance trends across varying turn lengths (6, 12, 18, 24) using diverse backbone models (Qwen3-8B, Qwen3-235B, Deepseek-v3.2) on the \textsc{CoSER} benchmark.

\subsubsection{Preliminaries}
To diagnose how a multi-turn role-playing agent reallocates reliance across competing context sources as the interaction horizon grows, we extract decoding-time \emph{attention/activation} signals rather than relying solely on end-task performance curves~\cite{jain-wallace-2019-attention, chefer-etal-2021-beyond-attention}. Concretely, for each generation, we partition the available prompt context into disjoint token segments (e.g., \textsc{Profile}, \textsc{History}, and \textsc{Generated}) and analyze the model’s attention at the \emph{final generated token}. Let $L$ be the number of transformer layers and $H$ the number of attention heads. At layer $\{\ell\in{1,\dots,L}\}$, we take the attention distribution from the last query position, average it across heads, and obtain $\bar{\alpha}^{(\ell)}\in[0,1]^{T}$ over the $T$ key/value positions. For a segment $s\subseteq{\{1,\dots,T}\}$ with token length $|s|$, the segment attention mass at layer $\ell$ is
\[
A_s^{(\ell)} = \sum_{i\in s} \bar{\alpha}^{(\ell)}_i
\]

\subsubsection{Results and Analysis}
\begin{figure}[h]
    \centering
    \includegraphics[width=0.9\linewidth]{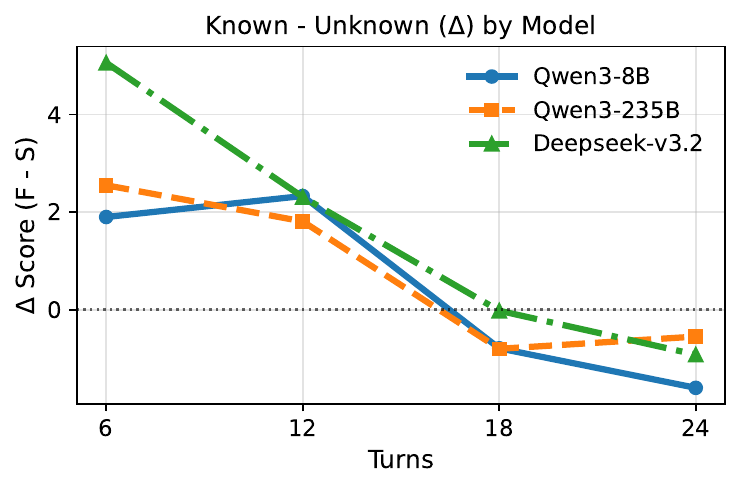}
    \vspace{-2mm}
    \caption{Turn-wise Trends in Role-Playing Performance of the Famous and Synthetic Groups}
    \vspace{-5mm}
    \label{fig:turn_ablation}
\end{figure}

Figure~\ref{fig:turn_ablation} plots the performance gap ($F - S$) across turn counts, revealing a consistent decline as interactions lengthen. This convergence results from the steady degradation of \textit{Known} characters contrasted with the robustness of \textit{Unknown} characters, which often improve over time. Notably, Deepseek-v3.2 exhibits a distinctive, unwavering decline as the turn count increases. Ultimately, extended interactions dilute the parametric advantage of \textit{Known} characters while allowing models to better simulate \textit{Unknown} characters through accumulated context.

\subsection{Attention and Saturation Analysis}
\subsubsection{Motivation}
As contexts lengthen, token-level cues from \textbf{profile} and \textbf{history context} become increasingly diluted and compete with the model’s own \textbf{self-conditioning} signal, potentially shifting generation away from explicit context grounding and toward stronger reliance on \textit{parametric knowledge}. Because this transition is \textit{not} directly observable from outputs alone, we complement turn ablations with mechanistic analysis during decoding~\cite{jain-wallace-2019-attention, chefer-etal-2021-beyond-attention}. We instrument Qwen3-8B at fixed horizons (Turn 6/12/18/24) under \textit{Known/Unknown} conditions by partitioning the context into \textbf{Profile}, \textbf{History}, and \textbf{Self-generated} segments. We quantify segment reliance using (i) attention lift at the final generated token (length-normalized attention preference)~\cite{abnar-zuidema-2020-attention-flow} and (ii) a saturation layer statistic capturing when each segment’s influence emerges across layers~\cite{tenney-etal-2019-bert-pipeline}. Together, these metrics succinctly characterize turn-length–dependent shifts in grounding versus self-conditioning. 

\subsubsection{Attention lift}
Raw attention mass $A_s^{(\ell)}$ is length-biased (longer segments accrue more mass under near-uniform attention). We therefore report \emph{attention lift} as a length-normalized preference relative to a uniform baseline over tokens:
\[
\mathrm{Lift}_s^{(\ell)} = \frac{A_s^{(\ell)}}{|s|/T} = A_s^{(\ell)}\cdot\frac{T}{|s|}
\]
Intuitively, $\mathrm{Lift}_s^{(\ell)}>1$ indicates that the model attends to segment $s$ more than expected from its token length, while $<1$ indicates under-attention. In our setting, we primarily use the final-layer value $\mathrm{Lift}_s^{(L)}$. This construction is closely related to attention-based attribution/rollout and flow-style analyses that aggregate and interpret attention signals across model components~\cite{abnar-zuidema-2020-attention-flow, chefer-etal-2021-beyond-attention}.

\paragraph{Saturation layer}
To summarize \emph{when} a segment begins to dominate computation across depth, we define a saturation-layer statistic using the layer-wise masses $A_s^{(\ell)}$. Let $C_s^{(\ell)}=\sum_{j=1}^{\ell}A_s^{(j)}$ be the cumulative mass up to layer $\ell$. Given a fixed threshold $\tau$ (we use $\tau=0.95$), the saturation layer is
\[
\mathrm{Sat}_s = {\arg\min}_{\ell \in \{1,\dots,L\}} \ \mathbb{I}\!\left[C_s^{(\ell)} \ge \tau\, C_s^{(L)}\right]
\]

Lower $\mathrm{Sat}_s$ means that segment $s$ exerts most of its cumulative influence early in the network, while higher values indicate later integration. This notion aligns with prior evidence that representations and decision-relevant information emerge progressively across transformer depth, motivating layer-wise diagnostics for interpretability~\cite{tenney-etal-2019-bert-pipeline, abnar-zuidema-2020-attention-flow}.

\paragraph{Efficiency notes}
Computing these metrics during autoregressive decoding is expensive for long prompts. We therefore use standard \emph{KV-caching}~\cite{sheng2023flexgen} to avoid recomputing attention for already-processed prefixes, and we apply a \emph{profile-prefill} strategy~\cite{10.1145/3759441.3759444} that incrementally pre-encodes the long system-profile region in fixed-size chunks (in our setting, 512 tokens) before stepwise decoding. These optimizations affect runtime and memory footprint but do not change the metric definitions.

\subsubsection{Results and Analysis}


\begin{table}[!htbp]
\centering
\small
\setlength{\tabcolsep}{3.5pt}
\renewcommand{\arraystretch}{1.15}
\resizebox{0.45\textwidth}{!}{
\begin{tabular}{ll|cccc|c}
\toprule
\multirow{2}{*}{\textbf{Metric}} & \multirow{2}{*}{\textbf{Group}} & \multicolumn{5}{c}{\textbf{Turns}} \\
\cmidrule(lr){3-7}
 & & \textit{T6} & \textit{T12} & \textit{T18} & \textit{T24} & $\Delta$\\
\midrule
\multirow{2}{*}{\textit{P. Lift}}
 & Known    & 0.68 & 0.70 & 0.74 & 0.75 & +0.07 \\
 & Unknown & 0.67 & 0.68 & 0.70 & 0.72 & +0.05 \\
\midrule
\multirow{2}{*}{\textit{H. Lift}}
 & Known    & 3.57 & 2.33 & 1.82 & 1.78 & -1.79 \\
 & Unknown & 2.87 & 1.95 & 1.58 & 1.43 & \textbf{-1.44} \\
\midrule
\multirow{2}{*}{\textit{G. Lift}}
 & Known    & 65.06 & 67.34 & 68.61 & 72.46 & +7.40 \\
 & Unknown & \textbf{56.36} & \textbf{58.75} & \textbf{61.88} &\textbf{ 63.76} & +7.40 \\
\midrule
\multirow{2}{*}{\textit{P. Satur}}
 & Known    & 6.55 & 7.42 & 7.92 & 8.34 & +1.79 \\
 & Unknown & \textbf{5.95} & \textbf{6.60} & \textbf{7.16} & \textbf{7.28} & \textbf{+1.33} \\
\midrule
\multirow{2}{*}{\textit{H. Satur}}
 & Known    & 0.59 & 0.33 & 0.25 & 0.22 & -0.37 \\
 & Unknown & 0.46 & 0.25 & 0.19 & 0.16 & -0.30 \\
\bottomrule
\end{tabular}
}
\caption{Turn-wise comparison of lift\_last and saturation between Known and Unknown groups.}
\label{tab:lift_saturation_famous_synth}
\vspace{-5mm}
\end{table}

Table~\ref{tab:lift_saturation_famous_synth} shows that longer horizons shift decoding away from explicit context grounding and toward self-conditioned, parametric generation. In both groups, \textit{H.~Lift} drops substantially with turns, while \textit{G.~Lift} increases, indicating reduced reliance on dialogue context and stronger dependence on the model’s own evolving output. The key group difference is magnitude: \textit{Known} maintains consistently higher \textit{G.~Lift} than \textit{Unknown} at every horizon, suggesting a more parametric/self-conditioned regime. Consistently, \textit{P.~Satur} rises in both settings but is higher and grows more in \textit{Known} (+1.79) than in \textit{Unknown} (+1.33), implying that persona constraints are incorporated later in the network for \textit{Known}more as late-stage correction than early conditioning. Taken together, the concurrent decrease in \textit{H.~Lift}, increase in \textit{G.~Lift}, and the higher \textit{P.~Satur} in the \textit{Known} condition indicate a turn-length–driven shift toward a more self-conditioned, parametric regime in which persona constraints from the provided profile are incorporated later and less proactively than in the unseen \textit{Unknown} condition. This suggests that in multi-turn scenarios, the parametric identity of \textit{Known} characters actually exerts a detrimental effect on role-playing performance.

\section{Detailed FACD Setup}
\label{app:facd_details}

FACD constructs the negative prompt by selectively removing profile fields classified as immoral while preserving the remaining character information as a sanitized behavioral baseline. Importantly, this procedure does not discard an entire immoral character profile. Instead, it operates at the field level, so that only specific fields whose content is classified as immoral are removed from the negative prompt.

\paragraph{Immoral-flagged example.}
As a concrete case study, consider the \textit{Immoral} character James Moriarty. The \texttt{Motivations.Goal} field is classified as immoral:

\begin{quote}
"To build and maintain an unassailable criminal empire spanning all of London and beyond; to exercise absolute control over the criminal underworld while remaining invisible to law enforcement; to prove his intellectual superiority; to accumulate wealth and power; to eliminate any threat to his organization, particularly his intellectual equal."
\end{quote}

This field directly encodes criminal intent, domination, and elimination of threats, and is therefore removed from the sanitized negative prompt. In contrast, the \texttt{Abilities.Emotional Abilities. Ability to regulate emotions} field is classified as moral:

\begin{quote}
"Exceptional emotional control; can suppress emotional responses in favor of calculated action; maintains composure in extreme situations; channels emotions into productive planning rather than impulsive action; the oscillating head movement may be an unconscious release of internal tension."
\end{quote}

Although this ability may support Moriarty's immoral behavior in context, the field itself describes emotional regulation rather than an immoral value, goal, or worldview. It is therefore retained in the negative prompt. This example illustrates that FACD removes immoral field content rather than broadly suppressing all information associated with an immoral character.

\paragraph{Disposition-robust fallback fields.}
When the immoral-field filtering leaves fewer than six non-\texttt{Personal Attributes} fields, FACD supplements the negative prompt with disposition-robust fallback fields. These fallback fields are selected from profile fields that are identified as disposition-robust in Figure~\ref{fig:rq2_Coser}. Specifically, the disposition-robust fields are \textit{Extraversion}, \textit{Neuroticism}, \textit{Openness} in \texttt{BFI}, \textit{Positive} in \texttt{Traits}, \textit{In\_Conflict\_Situations} in \texttt{Social\_Interaction}, all fields in \texttt{Relationships}, all fields in \texttt{Knowledge\_and\_Skills}. These fields are used to construct the fallback set $\mathcal{F}_{\mathrm{pad}}$.

As a result, the final negative prompt contains no field that is both disposition-sensitive and immoral. It instead serves as a coherent, sanitized baseline against which the full positive profile can be contrasted. The contrastive term in FACD, therefore, amplifies the contribution of the omitted immoral fields, especially when those fields occur in disposition-sensitive parts of the profile.


\begin{table*}[t]
\centering
\scalebox{0.9}{%
\begin{tabularx}{\textwidth}{X}
\toprule
\textbf{Scenario Generation Prompt} \\
\midrule
\textless System\textgreater

You are a narrative designer.

You are given structured profiles for several characters from the same fictional work.

Based on the information in the profiles below, write three discriminative scene descriptions, WITHOUT any dialogue lines.

Requirements:

- Each scenario should be 3 to 6 sentences.

- Focus on mood, recent or ongoing events, and how their goals, relationships, and emotions shape the situation.

- Do not invent completely new backstory that contradicts the profiles.

- Output only the scene description as plain text (no bullet points, no JSON).

\textless User\textgreater

Character profiles:

\{profiles\}
\\
\bottomrule
\end{tabularx}%
}
\caption{Prompt for Scenario Generation.}
\vspace{-5mm}
\label{prompt:scenario_generation}
\end{table*}

\begin{table*}[t]
\centering
\scalebox{0.9}{%
\begin{tabularx}{\textwidth}{X}
\toprule
\textbf{Profile Schema (Part 1.)} \\
\midrule

\ttfamily\footnotesize
\begin{tabular}{@{}l@{}}
"Personal Attributes": \{ \\
\hspace*{2em}"type": "object", \\
\hspace*{2em}"desc": "Individual's basic intrinsic traits.", \\
\hspace*{2em}"properties": \{ \\
\hspace*{4em}"Name": \{"type": "string", "desc": "Character's full name as stated in the work."\}, \\
\hspace*{4em}"Age": \{"type": "string", "desc": "Age as presented in the work (range or specific)."\}, \\
\hspace*{4em}"Gender": \{"type": "string", "desc": "male or female (only if explicit)."\}, \\
\hspace*{4em}"Origin": \{"type": "string", "desc": "Place of origin or nationality."\}, \\
\hspace*{4em}"Appearance": \{"type": "string", "desc": "Hair, build, attire, notable traits."\}, \\

\hspace*{2em}\}, \\
\}, \\
"Personality Traits": \{ \\
\hspace*{2em}"type": "object", \\
\hspace*{2em}"desc": "Individual's character, behavior, thoughts, feelings, etc.", \\
\hspace*{2em}"properties": \{ \\
\hspace*{4em}"Big5": \{ \\
\hspace*{6em}"type": "object", \\
\hspace*{6em}"desc": "Five-factor personality (short phrases).", \\
\hspace*{6em}"properties": \{ \\
\hspace*{8em}"Extraversion": \{"type": "string", "desc": "Sociability/energy."\}, \\
\hspace*{8em}"Conscientiousness": \{"type": "string", "desc": "Organization/discipline."\}, \\
\hspace*{8em}"Agreeableness": \{"type": "string", "desc": "Cooperativeness/empathy."\}, \\
\hspace*{8em}"Neuroticism": \{"type": "string", "desc": "Stability/reactivity."\}, \\
\hspace*{8em}"Openness": \{"type": "string", "desc": "Curiosity/creativity."\}, \\
\hspace*{6em}\}, \\
\hspace*{4em}\}, \\
\hspace*{4em}"Preference": \{ \\
\hspace*{6em}"type": "object", \\
\hspace*{6em}"desc": "Personal preferences.", \\
\hspace*{6em}"properties": \{ \\
\hspace*{8em}"Like": \{"type": "string", "desc": "Things liked/favored."\}, \\
\hspace*{8em}"Hate": \{"type": "string", "desc": "Things disliked/avoided."\}, \\
\hspace*{6em}\}, \\
\hspace*{4em}\}, \\
\hspace*{4em}"Character": \{ \\
\hspace*{6em}"type": "object", \\
\hspace*{6em}"desc": "Positive and negative traits.", \\
\hspace*{6em}"properties": \{ \\
\hspace*{8em}"Positive Traits": \{"type": "string", "desc": "Positive traits."\}, \\
\hspace*{8em}"Negative Traits": \{"type": "string", "desc": "Negative traits."\}, \\
\hspace*{6em}\}, \\
\hspace*{4em}\}, \\
\hspace*{2em}\}, \\
\}, \\
\end{tabular}

\\
\bottomrule
\end{tabularx}
}
\caption{Profile Schema Template (part 1).}
\label{prompt:template1}
\end{table*}

\begin{table*}[t]
\centering
\scalebox{0.9}{%
\begin{tabularx}{\textwidth}{X}
\toprule
\textbf{Profile Schema (part 2} \\
\midrule

\ttfamily\footnotesize
\begin{tabular}{@{}l@{}}
"Interpersonal Relationships": \{ \\
\hspace*{2em}"type": "object", \\
\hspace*{2em}"desc": "The dynamics of individual interactions within social contexts.", \\
\hspace*{2em}"properties": \{ \\
\hspace*{4em}"Social Interaction": \{ \\
\hspace*{6em}"type": "object", \\
\hspace*{6em}"desc": "Social tendencies by context.", \\
\hspace*{6em}"properties": \{ \\
\hspace*{8em}"In normal situations": \{"type": "string", "desc": "Usual behavior."\}, \\
\hspace*{8em}"In close relationships": \{"type": "string", "desc": "With intimates/allies."\}, \\
\hspace*{8em}"In conflict situations": \{"type": "string", "desc": "In disputes/opposition."\}, \\
\hspace*{6em}\}, \\
\hspace*{4em}\}, \\
\hspace*{4em}"Relationships": \{ \\
\hspace*{6em}"type": "object", \\
\hspace*{6em}"desc": "Relationships by polarity.", \\
\hspace*{6em}"properties": \{ \\
\hspace*{8em}"Positive Relationships": \{"type": "array\textless string\textgreater", "desc": "Friendly/allied and the ties."\}, \\
\hspace*{8em}"Negative Relationships": \{"type": "array\textless string\textgreater", "desc": "Hostile/strained and the ties."\}, \\
\hspace*{8em}"Neutral Relationships": \{"type": "array\textless string\textgreater", "desc": "Acquaintances/unspecified."\}, \\
\hspace*{6em}\}, \\
\hspace*{4em}\}, \\
\hspace*{2em}\}, \\
\}, \\
"Motivations": \{ \\
\hspace*{2em}"type": "object", \\
\hspace*{2em}"desc": "Prompts individuals to take action and determine their choices within specific contexts.", \\
\hspace*{2em}"properties": \{ \\
\hspace*{4em}"Goal": \{"type": "string", "desc": "Ultimate aims/motivations."\}, \\
\hspace*{4em}"Morality": \{"type": "string", "desc": "Moral standards/principles."\}, \\
\hspace*{4em}"Worldview": \{"type": "string", "desc": "Brief worldview of the world where the character lives."\}, \\
\hspace*{2em}\}, \\
\}, \\
"Abilities": \{ \\
\hspace*{2em}"type": "object", \\
\hspace*{2em}"desc": "Individual's proficiencies within specific domains.", \\
\hspace*{2em}"properties": \{ \\
\hspace*{4em}"Knowledge and Skills": \{ \\
\hspace*{6em}"type": "object", \\
\hspace*{6em}"desc": "Individual's grasp on domain-specific knowledge, technical skills.", \\
\hspace*{6em}"properties": \{ \\
\hspace*{8em}"Skills/Expertise": \{"type": "string", "desc": "Abilities or specialties."\}, \\
\hspace*{8em}"Education": \{"type": "string", "desc": "Education level/knowledge."\}, \\
\hspace*{6em}\}, \\
\hspace*{4em}\}, \\
\hspace*{4em}"Emotional Abilities": \{ \\
\hspace*{6em}"type": "object", \\
\hspace*{6em}"desc": "Emotional tendencies and self-awareness by context.", \\
\hspace*{6em}"properties": \{ \\
\hspace*{8em}"Commonly felt emotions": \{"type": "string", "desc": "Frequent emotions."\}, \\
\hspace*{8em}"Ability to regulate emotions": \{"type": "string", "desc": "Self-regulation capacity."\}, \\
\hspace*{8em}"Way of expressing emotions": \{"type": "string", "desc": "Expression style."\}, \\
\hspace*{6em}\}, \\
\hspace*{4em}\}, \\
\hspace*{2em}\}, \\
\}, \\
\end{tabular}

\\
\bottomrule
\end{tabularx}
}
\caption{Profile Schema Template (part 2).}
\label{prompt:template2}
\end{table*}

\begin{table*}[t]
\centering
\scalebox{0.9}{%
\begin{tabularx}{\textwidth}{X}
\toprule
\textbf{Scenario Generation Prompt} \\
\midrule
\textless System\textgreater

You are an expert analyst of fictional characters and a meticulous, canon-aware media scholar.

Task:
Given ONLY a character metadata, construct a complete hierarchical character profile that follows the provided schema exactly.  

Rules:

- Output a JSON object with EXACTLY five top-level keys:
  
  "Personal Attributes", "Personality Traits", "Interpersonal Relationships", "Motivations", "Abilities"
  
- Preserve the given Name EXACTLY at: "Personal Attributes.Name"

- If the character clearly matches a well-known fictional character, align details with established canon.

- If the name is ambiguous/unrecognized, produce a self-consistent, psychologically plausible profile.

- Keep all five top-level fields mutually compatible with each other.

- A work title may be provided ONLY as a disambiguation hint. Do NOT mention the work title anywhere in the output JSON.

- Output JSON only. No markdown. No commentary.

\textless User\textgreater

Fill ONLY the fields that this chunk supports or improves, as a JSON object (no markdown).
    
Allowed keys and brief descriptions: 

\{profile\_schema\}

Here is the metadata chunk: Summarize it at once.
    
\{metadata\}
\\
\bottomrule
\end{tabularx}
}
\caption{Prompt for Profile Summarization.}
\label{prompt:summarization}
\end{table*}

\begin{table*}[t]
\centering
\scalebox{0.9}{ 
\begin{tabularx}{\textwidth}{X}
\toprule
\textbf{Adopted Works (\# of Characters} \\
\midrule
\textless Movies\textgreater

Avatar (3),
DC (3),
Disney (3),
Harry Potter (5),
James Bond (3),
Marvel (5),
Game of Thrones (4),
The Matrix (3),
Transformers (3),
Resident Evil (3),
Star Wars (3),

\textless TV Shows\textgreater

Breaking Bad (3),
The Hunger Games (3),
Doctor Who (3),
Peaky Blinders (3),
SpongeBob (4),
Sherlock Holmes (3),
The Walking Dead (3),
The Lord of the Rings (3),
Star Trek (3),
Stranger Things (3),
The Witcher (3),

\textless Anime\textgreater

Attack on Titan (3),
Demon Slayer (3),
Fullmetal Alchemist (3),
Naruto (3),
Neon Genesis Evangelion (3),
One Piece (3),
Pokémon (3)

\textless Games\textgreater

Final Fantasy (3) ,
Persona 5 (3),
Pokémon (3),
The Legend of Zelda (3)

\\
\bottomrule
\end{tabularx}
}
\caption{Basis Works for Known Characters. The numbers in parentheses indicate the number of selected characters.}
\label{prompt:works}
\end{table*}

\begin{table*}[t]
\centering
\scalebox{0.9}{%
\begin{tabularx}{\textwidth}{X}
\toprule
\ttfamily\footnotesize
\begin{tabular}{@{}l@{}}
"profile": \{ \\
\hspace*{2em}"demographic information": \{ \\
\hspace*{4em}"age": \{"support": [21, 83], "sampling": "discrete set (10 values)"\}, \\
\hspace*{4em}"gender": \{"type": "categorical", "support": ["male", "female"]\}, \\
\hspace*{4em}"origin": \{"type": "categorical", "support": "25 locales "\}, \\
\hspace*{4em}"occupation": \{"type": "categorical", "support": "37 role categories"\}, \\
\hspace*{4em}"education": \{"type": "ordinal categorical", "support": "13 levels "\}, \\
\hspace*{4em}"residence": \{"type": "categorical", "support": "13 housing types "\}, \\
\hspace*{4em}"family\_status": \{"type": "categorical", "support": "15 household/relationship states"\}, \\
\hspace*{4em}"socioeconomic\_status": \{"type": "ordinal categorical", "support": "6 strata"\}, \\
\hspace*{4em}"health\_status": \{"type": "categorical", "support": "11 states "\} \\
\hspace*{2em}\}, \\
\hspace*{2em}"personality traits seed": \{"type": "categorical", "support": "30 natural-language descriptions"\}, \\
\hspace*{2em}"scenario genres": \{ \\
\hspace*{4em}"realistic\_contemporary": \{"support": "2 settings"\}, \\
\hspace*{4em}"scifi": \{"support": "10 settings"\}, \\
\hspace*{4em}"fantasy": \{"support": "10 settings"\}, \\
\hspace*{4em}"horror\_thriller": \{"support": "7 settings"\}, \\
\hspace*{4em}"specialized": \{"support": "13 settings"\}, \\
\hspace*{4em}"hybrid": \{"support": "7 settings"\} \\
\hspace*{2em}\} \\
\} \\
\end{tabular}

\\
\bottomrule
\end{tabularx}
}
\caption{Unknown Character's Skeleton Template.}
\label{prompt:synthetic sample}
\end{table*}

\begin{table*}[t]
\centering
\scalebox{0.9}{ 
\begin{tabularx}{\textwidth}{X}
\toprule
\textbf{Profile Coherence Validation Prompt} \\
\midrule
<System>
You are a profile coherence validator.
\\
Your task is to evaluate whether the given character profile contains any internal logical conflicts.
Assess coherence strictly in terms of internal consistency among attributes
(e.g., age–education–occupation, family\_status–age, health\_status–residence, genre–demographics).
Do not generate or imagine any scenario, story, or events.
\\
Requirements:
\\
- Judge only whether the profile is internally coherent.
- Assign a coherence score from 1 (severely inconsistent) to 10 (fully coherent).
- Identify whether the profile is valid or invalid based on logical consistency.
- Minor tensions are acceptable; mark invalid only if there are clear logical contradictions.
- Do not rely on cultural norms, stereotypes, or moral judgments.
- Do not invent missing information.

Allowed field names:
age, gender, origin, occupation, education, residence, family\_status,
socioeconomic\_status, health\_status, genre

Output format (JSON only):

```json\\
\hspace*{2em}\{\\
\hspace*{4em}  "is\_valid": true/false,\\
\hspace*{4em}  "coherence\_score": 1-10,\\
\hspace*{4em}  "problematic\_fields": ["field names causing logical issues"],\\
\hspace*{4em}  "issues": ["clear description of each inconsistency"],\\
\hspace*{4em}  "reasoning": "brief logic-based explanation"\\
\hspace*{2em}\}\\
\bottomrule
\end{tabularx}
}
\caption{Prompt for Profile Coherence Validation.}
\label{prompt:character Validation}
\end{table*}

\begin{table*}[t]
\centering
\scalebox{0.9}{ 
\begin{tabularx}{\textwidth}{X}
\toprule
\textbf{Story Generation Prompt} \\
\midrule
\textless System\textgreater
Write a compelling story (30,000 to 40,000 characters) featuring these three characters:

\#\# Character 1\\
\{persona1\}

\#\# Character 2\\
\{persona2\}

\#\# Character 3\\
\{persona3\}

\#\# Genre/Setting\\
\{genre\}
\\
\#\# Requirements:\\
- Show each character's personality through actions/dialogue, not exposition
\\
- Setting should feel authentic to the genre
\\
- Include meaningful conflict challenging the characters
\\
- Characters experience growth or revelation
\\
- All three characters should interact and have significant roles
\\
\\
Write the complete story. No meta-commentary—only the story.
\\
\bottomrule
\end{tabularx}
}
\caption{Prompt for Story Generation (Unknown Profile).}
\label{prompt:storyPrompt}
\end{table*}


\begin{table*}[t]
\centering
\scalebox{0.9}{%
\begin{tabularx}{\textwidth}{X}
\toprule
\textbf{Scenario Generation Prompt} \\
\midrule
\textless System\textgreater

You are a narrative designer.

You are given structured profiles for several characters from the same fictional work.

Based on the information in the profiles below, write three discriminative scene descriptions, WITHOUT any dialogue lines.

Requirements:

- Each scenario should be 3 to 6 sentences.

- Focus on mood, recent or ongoing events, and how their goals, relationships, and emotions shape the situation.

- Do not invent completely new backstory that contradicts the profiles.

- Output only the scene description as plain text (no bullet points, no JSON).

\textless User\textgreater

Character profiles:

\{profiles\}
\\
\bottomrule
\end{tabularx}%
}
\caption{Prompt for Scenario Generation.}
\vspace{-5mm}
\label{prompt:scenario_generation}
\end{table*}

\begin{table*}[t]
\centering
\scalebox{0.9}{
\begin{tabularx}{\textwidth}{@{}X@{}}
\toprule
\textbf{Human Evaluation Guideline} \\
\midrule

\textless Instruction\textgreater

Given the provided character profile and dialogue context, evaluate the
quality of the model response as a role-playing response.
Assign a single score from \textbf{1 to 5} by jointly considering
\textbf{Character Fidelity} and \textbf{Dialogue Quality}.
\\[4pt]

\textbf{\#\# Character Fidelity}

Evaluate how faithfully the response reflects the given character profile,
including the following aspects:

- \textbf{Language and Style}: Whether the vocabulary, expressions, tone,
and speaking style are appropriate for the character's personality and
social or educational background.

- \textbf{Knowledge and Background}: Whether the response appropriately
reflects the character's knowledge, background, and experiences without
introducing information the character should not know.

- \textbf{Personality and Behavior}: Whether the character's emotions,
actions, values, beliefs, interests, and decisions are consistent with
the provided profile.

- \textbf{Relationships and Social Status}: Whether interactions with other
characters are appropriate given the character's relationships, background,
and social position.
\\[4pt]

\textbf{\#\# Dialogue Quality}

Evaluate the overall coherence and narrative quality of the response,
including the following aspects:

- \textbf{Flow and Progression}: Whether the dialogue develops naturally
and makes meaningful progress.

- \textbf{Repetition}: Whether the response avoids unnecessary,
mechanical, or excessive repetition of previously stated information,
phrases, or perspectives.

- \textbf{Logical Consistency}: Whether the response remains logically
and factually consistent with the preceding dialogue and context.
\\[4pt]

\textbf{\#\# Scoring Guide}

\textbf{5 -- Excellent:}
The response is highly faithful to the character in both behavior and
language. The character's vocabulary, tone, knowledge, background, values,
decisions, and social interactions are consistently and naturally reflected.
The dialogue is coherent, progresses naturally, and contains no meaningful
contradictions or unnecessary repetition. Persona consistency remains strong
even in direct, sensitive, or provocative situations.
\\[3pt]

\textbf{4 -- Good:}
The response is generally faithful to the character and maintains an
appropriate character-specific style and behavior. Minor instances of
generic wording, slightly suboptimal decisions, or small inconsistencies
may occur, but they do not conflict with the character's core profile.
The dialogue remains coherent and natural overall.
\\[3pt]

\textbf{3 -- Fair:}
The response reflects the character only partially. The behavior is broadly
plausible but may be generic, insufficiently character-specific, or weakly
aligned with some aspects of the character's background, values, or
speaking style. Noticeable weaknesses in dialogue progression, repetition,
or consistency may appear, but there is no severe or pervasive violation
of the character profile.
\\[3pt]

\textbf{2 -- Poor:}
The response shows substantial misalignment with the character. Its behavior,
knowledge, values, relationships, or language frequently fail to reflect
the provided profile, and some parts may directly conflict with it.
The dialogue may also exhibit unnatural progression, excessive repetition,
or logical or factual inconsistencies.
\\[3pt]

\textbf{1 -- Very Poor:}
The response clearly fails to maintain the intended character. Core aspects
of the character's personality, background, knowledge, values, behavior,
relationships, or speaking style are absent or directly contradicted.
The dialogue may be incoherent, highly repetitive, logically inconsistent,
or otherwise implausible in context.
\\[4pt]

\textbf{\#\# Important}

Do \textbf{not} reward or penalize a response simply because the character
expresses morally desirable or undesirable attitudes.
Moral valence itself is not an evaluation criterion.
Evaluate whether the response faithfully and coherently reflects the
provided character profile and dialogue context.
\\[4pt]

\textbf{\#\# Output}

Return a single integer score from \textbf{1} to \textbf{5}.
\\

\bottomrule
\end{tabularx}
}
\caption{
Human evaluation prompt for role-playing quality.
Annotators assign a single 1--5 Likert score by jointly considering
character fidelity and dialogue quality.
}
\label{prompt:human_eval}
\end{table*}
\begin{table*}[t]
\centering
\scalebox{0.9}{ 
\begin{tabularx}{\textwidth}{X}
\toprule
\textbf{Profile Construction Prompt} \\
\midrule
\textless System\textgreater

You are an expert disposition scorer.

(Except BFI)

Task:

Given a single field from a JSON character profile, assign a score from 1 to 10:

- 1 = very immoral in terms of the given field

...

- 10 = very moral in terms of the given field

Rules (IMPORTANT):

- Use ONLY the provided field name and field content. Do NOT use outside knowledge.

- Think carefully and consider nuances, but DO NOT output any explanation.

- Your final output MUST be a single integer from 1 to 10, with no other text.

If the field is missing/empty/None or the evidence is unclear, choose a conservative score around 4 to 6.

Output score (STRICT):

(BFI)

Task:

Given a single field from a JSON character profile, assign a score from 1 to 10:

- 1 = very low in terms of the given field

...

- 10 = very high in terms of the given field

Rules (IMPORTANT):

- Use ONLY the provided field name and field content. Do NOT use outside knowledge.

- Think carefully and consider nuances, but DO NOT output any explanation.

- Your final output MUST be a single integer from 1 to 10, with no other text.

If the field is missing/empty/None or the evidence is unclear, choose a conservative score around 4 to 6.

Output score (STRICT): 

\textless User\textgreater

Field: \{field\}
\\
\bottomrule 
\end{tabularx}
}
\caption{Prompt for Judging Attributive Identity Type.}
\label{prompt:attributive_judge}
\end{table*}

\begin{table*}[t]
\centering
\scalebox{0.9}{%
\begin{tabularx}{\textwidth}{X}
\toprule
\textbf{Profile Construction Prompt} \\
\midrule
\textless System\textgreater \\
You are a character profile writer. You will be given a structured JSON profile of a fictional character.
Your task is to convert it into a single, cohesive unstructured prose description. 
Write in a descriptive, narrative style --- not bullet points or lists. 
Present the information naturally, as if describing the character to someone who has never encountered them. 
Cover all aspects from the structured profile: personal attributes, personality traits, relationships, 
motivations, and abilities. Do not omit any information from the original profile. 
The output should be between 500 and 600 words. 
Return ONLY a valid JSON object with two keys: ``Name'' (the character's name) and ``Character Summary'' (the prose description). 
No markdown, no extra commentary. \\
\\
\textless User\textgreater \\
\\
Convert the following structured character profile into an unstructured prose description. \\
\\
Structured profile: \\
\\
\{profile\_json\} \\
\bottomrule 
\end{tabularx}%
}
\caption{Prompt for Generating Unstructured Character Profiles.} 
\label{prompt:unstructure}
\end{table*}
\begin{table*}[t]
    \centering
    \small
    \setlength{\tabcolsep}{4pt}
    \renewcommand{\arraystretch}{1.1}
    \begin{tabular}{@{}l|ccc|ccc|ccc|ccc|ccc@{}}
    \toprule
    & \multicolumn{3}{c|}{\textbf{Qwen3-8B}}
    & \multicolumn{3}{c|}{\textbf{Qwen3-235B}}
    & \multicolumn{3}{c|}{\textbf{Mistral-Small}}
    & \multicolumn{3}{c|}{\textbf{Olmo-3.1-32B}}
    & \multicolumn{3}{c}{\textbf{DeepSeek-v3.2}} \\
    & K & U & $\Delta$
    & K & U & $\Delta$
    & K & U & $\Delta$
    & K & U & $\Delta$
    & K & U & $\Delta$ \\
    \midrule
    \rowcolor{gray!30}
    \multicolumn{16}{c}{\textbf{PersonaGym (Single-turn interview)}} \\
    \midrule
    \textit{Expected Action} & 4.65 & 4.64 & -0.01 & 4.81 & 4.78 & -0.03 & 4.76 & 4.73 & -0.02 & 4.74 & 4.78 & +0.04 & 4.62 & 4.65 & +0.03 \\
    \textit{Toxicity} & 3.99 & 3.87 & -0.12 & 4.23 & 4.08 & -0.15 & 4.27 & 4.27 & -0.00 & 4.78 & 4.79 & +0.01 & 4.26 & 4.28 & +0.02 \\
    \textit{Linguistic Habits} & 4.46 & 4.34 & -0.12 & 4.64 & 4.66 & +0.02 & 4.45 & 4.36 & -0.09 & 4.38 & 4.51 & +0.13 & 4.49 & 4.54 & +0.04 \\
    \textit{Persona Consistency} & 4.67 & 4.67 & +0.01 & 4.90 & 4.91 & +0.01 & 4.72 & 4.65 & -0.06$^{}$ & 4.74 & 4.68 & -0.06 & 4.72 & 4.70 & -0.03 \\
    \textit{Action Justification} & 4.71 & 4.63 & -0.08$^{}$ & 4.71 & 4.64 & -0.08$^{}$ & 4.70 & 4.62 & -0.08$^{}$ & 4.78 & 4.64 & -0.14 & 4.54 & 4.53 & -0.01 \\
    \rowcolor{gray!20}
    \textit{PersonaScore} & 4.50 & 4.43 & -0.06$^{}$ & 4.66 & 4.61 & -0.04 & 4.58 & 4.53 & -0.05$^{}$ & 4.68 & 4.68 & 0.00 & 4.53 & 4.54 & +0.01 \\
    \midrule
    \rowcolor{gray!30}
    \multicolumn{16}{c}{\textbf{CoSER (Multi-turn interaction)}} \\
    \midrule
    \textit{Anthropomorphism} & 11.00 & 12.44 & +1.44 & 24.54 & 27.68 & +3.14 & 17.29 & 21.00 & +3.71 & 10.45 & 12.98 & +2.53 & 26.15 & 29.30 & +3.15 \\
    \textit{Character Fidelity} & 7.42 & 10.65 & +3.23 & 18.82 & 18.81 & -0.01 & 20.07 & 23.61 & +3.54 & 12.47 & 13.86 & +1.39 & 27.94 & 31.00 & +2.06 \\
    \textit{Storyline Quality} & 43.61 & 47.24 & +3.63 & 56.52 & 58.52 & +2.00 & 54.53 & 57.29 & +2.76 & 45.41 & 47.49 & +2.08 & 63.20 & 64.58 & +1.38 \\
    \rowcolor{gray!20}
    \textit{avg.} & 20.67 & 23.44 & +3.23 & 33.30 & 35.00 & +1.70 & 30.63 & 33.97 & +3.34 & 22.77 & 24.78 & +2.01 & 39.10 & 41.63 & +2.53 \\
    \bottomrule
    \end{tabular}
    \caption{Per-metric evaluation comparing Familiarity axis (Known (K) vs.\ Unknown (U)) across models. $\Delta = \text{K} - \text{U}$.}
    \label{tab:known_vs_unknown}
\end{table*}

\begin{table*}[t]
    \centering
    \small
    \setlength{\tabcolsep}{4pt}
    \renewcommand{\arraystretch}{1.1}
    \begin{tabular}{@{}l|ccc|ccc|ccc|ccc|ccc@{}}
    \toprule
     & \multicolumn{3}{c|}{\textbf{Qwen3-8B}}
     & \multicolumn{3}{c|}{\textbf{Qwen3-235B}}
     & \multicolumn{3}{c|}{\textbf{Mistral-Small}}
     & \multicolumn{3}{c|}{\textbf{Olmo-3.1-32B}}
     & \multicolumn{3}{c}{\textbf{DeepSeek-v3.2}} \\
     & St & Un & $\Delta$
     & St & Un & $\Delta$
     & St & Un & $\Delta$
     & St & Un & $\Delta$
     & St & Un & $\Delta$ \\
    \midrule
    \rowcolor{gray!30}
    \multicolumn{16}{c}{\textbf{PersonaGym (Single-turn interview)}} \\
    \midrule
    \textit{Expected Action} & 4.64 & 4.62 & -0.03 & 4.80 & 4.76 & -0.04 & 4.75 & 4.74 & -0.00 & 4.76 & 4.72 & -0.04 & 4.63 & 4.65 & +0.01 \\
    \textit{Toxicity} & 3.93 & 4.05 & +0.11 & 4.16 & 4.25 & +0.09 & 4.27 & 4.19 & -0.08 & 4.78 & 4.50 & -0.28 & 4.27 & 4.30 & +0.03 \\
    \textit{Linguistic Habits} & 4.41 & 4.39 & -0.01 & 4.65 & 4.56 & -0.09 & 4.40 & 4.36 & -0.05 & 4.45 & 4.34 & -0.11 & 4.51 & 4.47 & -0.04 \\
    \textit{Persona Consistency} & 4.67 & 4.62 & -0.05 & 4.91 & 4.82 & -0.09 & 4.69 & 4.71 & +0.02 & 4.71 & 4.77 & +0.06 & 4.71 & 4.69 & -0.02 \\
    \textit{Action Justification} & 4.67 & 4.72 & +0.05 & 4.68 & 4.68 & +0.00 & 4.66 & 4.69 & +0.03 & 4.71 & 4.75 & +0.04 & 4.53 & 4.46 & -0.08 \\
    \rowcolor{gray!20}
    \textit{PersonaScore} & 4.46 & 4.48 & +0.01 & 4.64 & 4.61 & -0.03 & 4.55 & 4.54 & -0.01 & 4.68 & 4.62 & -0.07 & 4.53 & 4.51 & -0.02 \\
    \midrule
    \rowcolor{gray!30}
    \multicolumn{16}{c}{\textbf{CoSER (Multi-turn interaction)}} \\
    \midrule
    \textit{Anthropomorphism} & 11.72 & 14.57 & +2.85 & 26.11 & 25.39 & -0.72 & 19.14 & 20.85 & +1.71 & 11.86 & 11.72 & -0.14 & 27.73 & 32.13 & +4.40 \\
    \textit{Character Fidelity} & 9.04 & 11.31 & +2.27 & 18.82 & 15.94 & -2.88 & 21.84 & 22.88 & +1.04 & 13.11 & 13.63 & +0.52 & 29.47 & 24.35 & -5.12 \\
    \textit{Storyline Quality} & 45.42 & 41.23 & 4.19 & 57.52 & 53.66 & -3.86 & 55.91 & 57.80 & +1.89 & 46.52 & 47.79 & +1.27 & 63.89 & 61.83 & -2.06 \\
    \rowcolor{gray!20}
    \textit{avg.} & 22.06 & 22.37 & +0.31 & 34.15 & 31.66 & -2.49 & 32.30 & 33.84 & +1.54 & 23.55 & 24.38 & +0.83 & 40.36 & 39.44 & -0.92 \\
    \bottomrule
    \end{tabular}
    \caption{Per-metric evaluation comparing Structure (Structured (St) vs.\ Unstructured (Un)) axis across models. $\Delta = \text{St} - \text{Un}$.}
    \label{tab:struct_vs_unstruct}
\end{table*}

\begin{table*}[t]
    \centering
    \small
    \setlength{\tabcolsep}{4pt}
    \renewcommand{\arraystretch}{1.1}
    \begin{tabular}{@{}l|ccc|ccc|ccc|ccc|ccc@{}}
    \toprule
    & \multicolumn{3}{c|}{\textbf{Qwen3-8B}}
    & \multicolumn{3}{c|}{\textbf{Qwen3-235B}}
    & \multicolumn{3}{c|}{\textbf{Mistral-Small}}
    & \multicolumn{3}{c|}{\textbf{Olmo-3.1-32B}}
    & \multicolumn{3}{c}{\textbf{DeepSeek-v3.2}} \\
    & M & I & $\Delta$
    & M & I & $\Delta$
    & M & I & $\Delta$
    & M & I & $\Delta$
    & M & I & $\Delta$ \\
    \midrule
    \rowcolor{gray!30}
    \multicolumn{16}{c}{\textbf{PersonaGym (Single-turn interview)}} \\
    \midrule
    \textit{Expected Action} & 4.65 & 4.64 & -0.01 & 4.81 & 4.79 & -0.02 & 4.79 & 4.70 & -0.10$^{}$ & 4.74 & 4.88 & +0.14 & 4.63 & 4.64 & +0.01 \\
    \textit{Toxicity} & 4.87 & 2.99 & -1.88$^{}$ & 4.89 & 3.41 & -1.48$^{}$ & 4.90 & 3.63 & -1.27$^{}$ & 4.99 & 3.77 & -1.22 & 4.93 & 3.61 & -1.32$^{}$ \\
    \textit{Linguistic Habits} & 4.38 & 4.44 & +0.06 & 4.67 & 4.63 & -0.04 & 4.42 & 4.39 & -0.02 & 4.43 & 4.55 & +0.12 & 4.54 & 4.49 & -0.05$^{}$ \\
    \textit{Persona Consistency} & 4.61 & 4.73 & +0.13$^{}$ & 4.93 & 4.89 & -0.04$^{}$ & 4.72 & 4.65 & -0.07$^{}$ & 4.72 & 4.65 & -0.07 & 4.71 & 4.71 & -0.01 \\
    \textit{Action Justification} & 4.70 & 4.63 & -0.07$^{}$ & 4.72 & 4.64 & -0.08$^{}$ & 4.72 & 4.60 & -0.13$^{}$ & 4.71 & 4.72 & +0.01 & 4.56 & 4.51 & -0.04 \\
    \rowcolor{gray!20}
    \textit{PersonaScore} & 4.64 & 4.29 & -0.36$^{}$ & 4.80 & 4.47 & -0.33$^{}$ & 4.71 & 4.39 & -0.32$^{}$ & 4.72 & 4.51 & -0.20 & 4.67 & 4.39 & -0.28$^{}$ \\
    \midrule
    \rowcolor{gray!30}
    \multicolumn{16}{c}{\textbf{CoSER (Multi-turn interaction)}} \\
    \midrule
    \textit{Anthropomorphism} & 14.53 & 8.55 & -5.98 & 29.14 & 22.63 & -6.51 & 20.27 & 17.85 & -2.42 & 15.47 & 10.56 & -7.91 & 29.46 & 25.78 & -3.68 \\
    \textit{Character Fidelity} & 11.39 & 6.39 & -5.00 & 23.53 & 13.41 & -10.12 & 27.12 & 15.78 & -11.3 & 15.24 & 7.58 & -7.66 & 32.78 & 25.73 & -7.05 \\
    \textit{Storyline Quality} & 48.72 & 41.71 & -7.01 & 61.51 & 52.95 & -8.56 & 57.74 & 53.82 & -3.92 & 49.71 & 42.72 & -6.99 & 66.22 & 61.28 & -4.94 \\
    \rowcolor{gray!20}
    \textit{avg.} & 24.88 & 18.88 & -6.00 & 38.06 & 29.66 & +8.40 & 35.04 & 29.15 & -5.89 & 26.81 & 20.28 & -6.53 & 42.82 & 37.60 & -5.22 \\
    \bottomrule
    \end{tabular}
    \caption{Per-metric evaluation comparing Disposition axis (Moral (M) vs.\ Immoral (I)) across models. $\Delta = \text{M} - \text{I}$.}
    \label{tab:moral_vs_immoral}
\end{table*}

\begin{table*}[t]
\centering
\scalebox{0.9}{ 
\begin{tabularx}{\textwidth}{>{\raggedright\arraybackslash}X}
\toprule
\textbf{Disposition Rewritting Prompt} \\
\midrule
\textless System\textgreater \newline
You rewrite a single field of a fictional character profile to flip its moral polarity. \newline
You will be told the target polarity: \newline
\hspace*{2ex}- MORAL: rewrite the field so the character comes across as ethical, compassionate, prosocial. \newline
\hspace*{2ex}- IMMORAL: rewrite the field so the character comes across as unethical, cruel, antisocial. \newline
Hard constraints (all must hold): \newline
\hspace*{2ex}1. Word count must be within $\pm$5\% of the original word count. \newline
\hspace*{2ex}2. Preserve specificity: keep the same level of concrete detail. If the original \newline
\hspace*{5ex}names entities, settings, skills, or events, the rewrite must include \newline
\hspace*{5ex}analogous specific elements (you may swap them for ones consistent with the \newline
\hspace*{5ex}new polarity, but do NOT generalize into vague language). \newline
\hspace*{2ex}3. Preserve the number of atomic facts (independent claims / list items). \newline
\hspace*{5ex}If the original is a comma- or semicolon-separated list of N items, output \newline
\hspace*{5ex}N items. If it has N independent clauses, output N independent clauses. \newline
\hspace*{2ex}4. Preserve writing style: same tone, register, sentence structure, punctuation \newline
\hspace*{5ex}pattern, and approximate sentence count as the original. \newline
\hspace*{2ex}5. Do not mention the character's name unless the original does. \newline
\hspace*{2ex}6. Output ONLY the rewritten field text. No preamble, no quotes, no explanation. \newline
\textless User\textgreater \newline
Target polarity: \{polarity\} \newline
Field path: \{path\} \newline
Original word count: \{wc\} \newline
Original field text: \newline
\textless\textless\textless \newline
\{text\} \newline
\textgreater\textgreater\textgreater \newline
Rewrite the field text following all constraints. Output the rewritten text only. \\
\bottomrule
\end{tabularx}
}
\caption{Prompt for Disposition Rewritting.}
\label{prompt:rewrite}
\end{table*}

\end{document}